\documentclass[11pt]{article}

\usepackage[final]{acl}

\usepackage{times}
\usepackage{latexsym}

\usepackage{xspace}

\usepackage[T1]{fontenc}

\usepackage[utf8]{inputenc}

\usepackage{microtype}

\usepackage{inconsolata}

\usepackage{graphicx}
\usepackage{tikz}
\usepackage{array}
\usepackage{enumitem}
\usepackage{makecell}
\usepackage{hyperref}
\usepackage{todonotes}

\usetikzlibrary{calc}
\usetikzlibrary{positioning}

\definecolor{darkyellow}{HTML}{9e7402} 
\newcommand{\ui}{\href{https://culture-in-nlp.pages.dev/}{\textsc{\textbf{\textcolor{darkyellow}{CultureMine}}}}\xspace}

\NewDocumentCommand\emojione{}{\raisebox{-0.75em}{\includegraphics[height=2.5em]{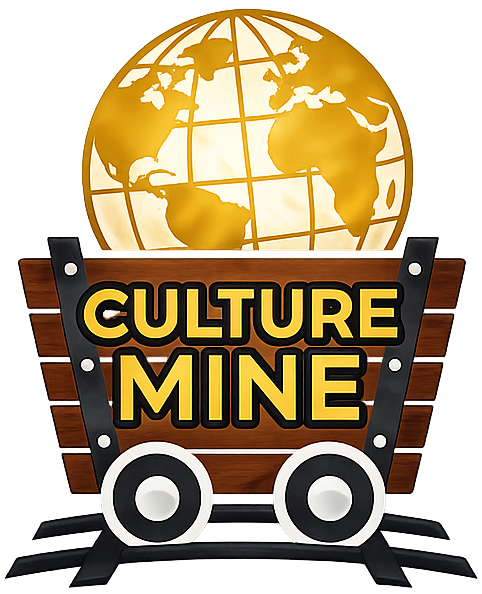}}}

\title{\emojione\hspace{0.1em} \ui: Datasets and Methods for Improving the Cultural Capabilities of NLP Systems}

\author{
  \textbf{Tania Chakraborty}\textsuperscript{$\spadesuit$}\thanks{Equal contribution.} \enspace
  \textbf{Eylon Caplan}\textsuperscript{$\spadesuit$*} \enspace
  \textbf{Zhaoqing Wu}\textsuperscript{$\spadesuit$*} \enspace
  \textbf{Kevin Cushing}\textsuperscript{$\spadesuit$} \enspace
  \textbf{Han Qin}\textsuperscript{$\spadesuit$} \enspace
\\
  \textbf{Shreya Havaldar}\textsuperscript{$\clubsuit$} \enspace
  \textbf{Dan Goldwasser}\textsuperscript{$\spadesuit$}
\\
  \textsuperscript{$\spadesuit$}Purdue University, \textsuperscript{$\clubsuit$}University of Pennsylvania
\\
  {\texttt{\{tchakrab,ecaplan,wu1828\}@purdue.edu}}
}

\begin{document}
\maketitle
\begin{abstract}
In recent years, there has been a surge of interest in Cultural NLP, with substantial efforts to create globally inclusive NLP systems. The rapid growth of literature in this field makes it difficult to track trends in methods and data resources. To address this, we analyze over 375 papers to answer three complementary questions: (1) What \emph{Cultural Capabilities (CCs)} are being targeted in NLP systems? (2) How are \emph{cultural data resources} being created? and (3) What \emph{methods} are being used to improve the CCs of those systems? We discuss trends observed across the three questions, and identify relevant research gaps. To facilitate further research in this field, we release our full list of analyzed papers, in the form of an interactive web interface, \ui\footnote{Accessible now at \href{https://culture-in-nlp.pages.dev/}{https://culture-in-nlp.pages.dev/}.}, which includes a feature to allow researchers to add their work; we hope this facilitates future research and proves to be a valuable resource for the Cultural NLP community.
\end{abstract}

\section{Introduction}

\begin{center}
\textit{“No people come into possession of a culture without having paid a heavy price for it.”}
\end{center}
\begin{flushright}
\textit{— James Baldwin}
\end{flushright}


\input{figures/top_right}

Language and culture are fundamentally interdependent; language both reflects, and is shaped by culture \cite{2598291a-d0b6-399c-a4fa-1f25e75dadc3}. Cultural NLP, which studies this intersection, has seen rapid growth in recent years; As of March 2026, searching the ACL Anthology for the word "culture" or "cultural" yields more than 700 papers from the last two years alone!
This volume of emergent work makes it difficult for researchers to keep abreast of novel methodologies and data resources for Cultural NLP.
To address this, we survey over 375 papers and provide an overview of (1) what Cultural Capabilities (CCs) are being targeted for improvement (\S\ref{sec:cultural-capabilities}), (2) methods utilized for \textbf{creating} cultural datasets (\S\ref{sec:data}), and (3) methods proposed to \textbf{improve} the CCs of NLP systems (\S\ref{sec:methods}).

To help navigate the volume of papers, we first group them by contribution type---\textit{datasets}, and \textit{systems}, and then each contribution is also mapped to a CC, based on what cultural competency it targets. Within each contribution type, we propose a taxonomy to organize papers, and answer the questions: (1) For dataset contributions, how are cultural datasets created? and (2) For system contributions, what methods are adopted to improve the CCs of NLP systems? In contrast to other surveys in this field (discussed in \S\ref{sec:related_works}), we offer a technical overview of the field and propose an organization scheme to navigate a voluminous and rapidly growing body of literature.

Furthermore, we release our full list of surveyed papers through an interactive web interface, \ui{}; each paper is manually annotated with our taxonomy---in addition to metadata such as cultural proxy (e.g., languages, countries). \ui{} also includes a submission form that researchers can use to add their work to the collection\footnote{Submissions to \ui{} are reviewed by the authors to ensure tag consistency.}.
We hope this will be a valuable resource for the cultural NLP community.

Our contributions are as follows:
\begin{itemize}
    \item We survey over 375 papers, organize them based on our proposed taxonomy, and additionally tag them with rich cultural metadata.
    \item We identify trends in methodologies to provide the reader with an overview of the field, and propose future directions for cultural NLP.
    \item We release \ui{} as a resource for the community. We welcome additions to \ui{} to foster collaborative research and streamline the discovery of relevant work.
\end{itemize}

\section{Related Surveys}
\label{sec:related_works}

There are some highly relevant surveys for cultural NLP, offering insights into definitions of culture and how culture is operationalized in NLP. \citet{liu2025culturally, adilazuarda2024measuringmodelingculturellms} discuss how culture is defined, and survey commonly adopted proxies of culture.
\citet{zhou2025culture} offer a nuanced perspective on how cultural NLP systems should be designed, based on theories from sociocultural linguistics.
In this paper, we don't redefine culture, and refer readers to the surveys mentioned above for this. To select papers we survey, we deferred to the authors to determine whether their work is cultural NLP or not (\S\ref{sec:lit_collection} for details).
Our survey adopts a complementary, technical perspective. Our main goal is to present an overview of recent \textbf{technical advances} for improving the cultural capabilities of NLP systems.

A closely related survey is \citet{pawar2025survey}, which surveys cultural awareness in language models, with a focus on data resources, and provides a comprehensive overview of resources for the community.
In contrast, we survey NLP systems broadly without limiting the survey to language models. Additionally, our focus is on the methods, both for creating data resources as well as improving NLP systems. Finally, more than half the papers in our survey are from 2025, and thus not included in \citet{pawar2025survey}.

\section{Literature Collection and Annotation}
\label{sec:lit_collection}
In this section we provide a brief overview of our literature collection and paper annotation strategy. Papers were added manually by authors as well as automatically via a keyword scrape of the ACL Anthology covering the past five years (up to and including 2025). After the initial keyword scrape, we utilized an LLM-assisted filtering pipeline to retain papers that met our criteria; they proposed either a novel methodology or dataset for cultural NLP.

All remaining papers were manually filtered by the authors to filter out works whose primary contributions were sociolinguistic or sociological rather than computational (see \S\ref{sec:contribution-categories} for survey scope). This resulted in a final corpus of over 375 papers, which were read and annotated by the authors. We held regular meetings to refine the definitions of our taxonomies; All papers were reviewed to ensure strict adherence to definitions in \S\ref{subsec:cc-tax}, \S\ref{sec:data}, and \S\ref{sec:methods}. Full details regarding our search queries, LLM filtering pipeline, and consensus protocol are provided in Appendix~\ref{app:collection}. In the subsequent sections, we cite only representative works, but a full list of papers can be found in Appendix~\ref{sec:allpapers}.
\section{Definitions}
\label{sec:cultural-capabilities}

In this section, we formally define the lens through which we analyze the surveyed literature. We define a traditional NLP \textit{task} as a well-defined computational problem specifying the behavior a system should exhibit. In contrast to standard tasks (e.g., classification, text generation), in this paper we categorize works based on the \textit{Cultural Capability} they address. The distinction is clarified and further explained in the following subsection.

\subsection{Defining Cultural Capabilities}
\label{subsec:cc-tax}
We define a \textbf{Cultural Capability (CC)} as an abstraction over NLP tasks to answer: \textit{``What specific cultural behavior is this dataset measuring, or is this system improving?''}

A CC may be measured via various NLP tasks, and a given NLP task can be used to measure various CCs. To illustrate, a system's Cultural Knowledge could be measured via the NLP task of QA (e.g., ``What foods are strictly prohibited in a kosher diet?''), but could also be measured via a recipe generation task, where outputting a dish with pork is inherently penalized as a factual error.

The motivation for focusing on CCs rather than on standard NLP tasks was made early in survey phase. It enabled us to identify what kinds of CCs are being targeted by the community, and which ones may be overlooked. Looking at contributions through the lens of CCs instead of NLP tasks allows for a more nuanced understanding of how culturally competent current NLP systems are.

The CCs were not pre-defined, but rather discovered through our bottom-up survey of the literature. We identified nine core CCs that the NLP community is actively working to measure and improve:

\input{figures/datasets}

\begin{enumerate}[leftmargin=*, nosep]\setlength\itemsep{0em}\setlength\topsep{0em}
    \item \textbf{Cultural Translation:} the ability to convert text in one language to another, while preserving/accounting for cultural nuance and artifacts.
    \item \textbf{Survey-based Cultural Alignment:} the capability of predicting the distributions of answers to value surveys conditioned on particular cultures (e.g., World Value Survey).
    \item \textbf{Value-driven Cultural Alignment:} the capability to behave in such a way that agrees with a particular culture's values, or to switch between behaviors aligning with target cultures.
    \item \textbf{Cultural Knowledge:} the ability to know, recall, understand, and use textual knowledge about a particular culture or cultures.
    \item \textbf{Multimodal Cultural Knowledge:} the ability to know, recall, understand, and use visual knowledge about a particular culture or cultures.
    \item \textbf{Sociolinguistic Competence:} the ability to understand, produce, or use language in such a way that it aligns with a particular culture or cultures' expected linguistic behavior.
    \item \textbf{Cultural Safety and Harm Reduction:} the ability to understand, censor, correct, or avoid text that contains harms or potential harms in accordance with a target culture or cultures.
    \item \textbf{Cultural Education:} the ability to aid in educative processes, conditioned on a target culture.
    \item \textbf{Computational Cultural Representation:} the capability to computationally represent, compare, modify, or edit representations of culture or its features.
\end{enumerate}

\subsection{Scope and Contribution Categorization}
\label{sec:contribution-categories}
\textbf{Survey Scope:} We limit the scope of this survey to include only works whose overarching aim is to improve or measure these CCs of NLP systems. As a result, we excluded works that use existing NLP models for computational social science (CSS). For example, \citet{garimella-etal-2016-identifying} use NLP methods to answer the question \emph{How are the same words used differently in different cultures?} This is valuable work but falls outside the scope of this paper. We view CSS as a vital pillar of Cultural NLP, but constrain our taxonomy specifically to those works that contribute a dataset or system advancement.

\textbf{Contribution Types:} Within the papers that meet our criteria, we first categorized them by their contribution type: \textbf{Datasets} and \textbf{Systems}. Note that many papers have both types of contributions. The main motivation for this categorization was to make it easier to analyze the broad field of Cultural NLP and extract insights about community focus and trends. Additionally, it aids researchers in efficiently finding resources via our companion webpage \ui{}. In \S\ref{sec:data}, we explore the Dataset branch, analyzing the methodologies used to curate cultural data. In \S\ref{sec:methods}, we analyze the Systems branch, detailing the NLP methods used for Evaluation, Data Generation, and Improvement.

\section{Datasets for Cultural Capabilities}

\label{sec:data}

We surveyed 320 dataset papers, and analyze \textit{techniques} and \textit{sources} for dataset creation, followed by insights into common dataset creation pipelines. The goal of this section is to leave the reader with an overview of \textit{how culture is being injected into datasets}.


\subsection{Dataset Creation Methods}
We broadly categorize cultural NLP dataset creation into four approaches: human-generated data (\S\ref{sec:humangenerated}), adaptation from existing language data sources (\S\ref{sec:languagesource}), datasets grounded in cultural research and frameworks (\S\ref{sec:cultureresearch}), and synthetic data generation (\S\ref{sec:syntheticdata}). Figure \ref{fig:datasets} shows our taxonomy.

\subsubsection{Human-Sourced Data} \label{sec:humangenerated}

\textbf{Crowdsourcing} workers are widely used for large-scale data collection and annotation requiring general human judgment rather than specialized cultural expertise: asking contributors to submit diverse images or questions \citep{cahyawijaya-etal-2025-crowdsource, arora-etal-2025-calmqa}, collecting opinions through voting \citep{falk-etal-2024-overview}, rating \citep{casola-etal-2024-multipico}, or abusive content labeling \citep{muhammad-etal-2025-afrihate}.\\
\textbf{Cultural experts or native speakers} are employed for tasks requiring cultural or linguistic expertise, such as labeling cultural aspects \citep{maji-etal-2025-drishtikon, alwajih-etal-2025-palm, xie-etal-2025-large}, creating culturally grounded content \citep{zhan-etal-2024-renovi, guo-etal-2025-care}, or assisting with translation \citep{kim-etal-2025-representing, montalan-etal-2025-batayan}.\\
\textbf{Domain experts} help ensure datasets align with the domains: law specialists developing annotation guidelines \citep{ullah-etal-2024-detecting}, historians and archaeologists ensuring artifact accuracy \citep{ghaboura-etal-2025-time}, and experts defining dataset taxonomies \citep{vasilev-etal-2025-ruscode}.

\subsubsection{Naturally Occurring Sources} \label{sec:languagesource}
\textbf{Web data} provides unstructured cultural signals in online reviews \citep{zou-etal-2025-llms} and social media to capture cross-cultural language use \citep{kumar-jurgens-2025-rules, wuraola-etal-2024-understanding, abdelkadir-etal-2024-diverse, kiesel-etal-2022-identifying, liu-etal-2025-mcs}, and images for visual QA and reasoning \citep{liu-etal-2021-visually, liu-etal-2025-mcs, bayramli-etal-2025-diffusion}.\\
\textbf{Parallel Corpora} contain comparable content across languages, such as culturally relevant entities \citep{yao-etal-2024-benchmarking, conia-etal-2024-towards}, rules \citep{haberland-etal-2024-italian}, and topical images \citep{schneider-sitaram-2024-m5}.\\
\textbf{Culture knowledge sources} offer explicit and fine-grained cultural concepts and artifacts, reflecting the highest level of culture sensitivity:
modifying existing datasets for a particular culture \cite{wang-etal-2024-kulture, grandury-etal-2025-la, son-etal-2025-kmmlu}, and filtering pre-existing cultural resources \cite{dai-etal-2025-word}.
Educational and domain-specific materials are also widely used: children's books \cite{khanuja-etal-2024-image}, exams \cite{cheng-etal-2025-hkcanto}, e-learning platforms \cite{pramodya-etal-2025-sinhalammlu}, research journals (e.g., PubMed) \cite{nimo-etal-2025-africa}, literature \cite{abuhaija-etal-2025-nakba}, and Wikipedia \cite{magdy-etal-2025-jawaher, bhatia-etal-2025-swan}.

\subsubsection{Cultural Research} \label{sec:cultureresearch}
\textbf{Culture Value Surveys}, such as the World Values Survey (WVS) \cite{inglehart2000world, haerpfer2022world}, provide standardized value-oriented questions and empirical responses that researchers use to construct datasets: expanding question sets \cite{xu-etal-2025-self}, predicting answer distributions \cite{cao-etal-2025-specializing}, or selecting dialogue topics based on response patterns \cite{ma-etal-2025-enhancing}.\\
\textbf{Cultural/Social Science Theories} also guide dataset creation. Frameworks such as Hofstede’s Cultural Dimensions \cite{hofstede1984culture}, the Theory of Basic Human Values \cite{SCHWARTZ19921}, and negotiation theory \cite{aslani_dignity_2016} are used to refine value taxonomies for question generation \cite{cahyawijaya-etal-2025-high}, categorize values \cite{yao-etal-2024-value}, and guide annotators \cite{hale-etal-2025-kodis}.

\subsubsection{Synthetic Generation} \label{sec:syntheticdata}
\textbf{Machine Translation} is used to expand datasets across languages. 
Work applies Google Translate to translate simple sentences \citep{belay-etal-2025-culemo, yin-etal-2022-geomlama} and literature \citep{thai-etal-2022-exploring}. 
LLMs are used for translation \citep{onohara-etal-2025-jmmmu, kim-kim-2025-dual}, augmentation \citep{masala_vorbesti_2024}, and verification of cultural alignment \citep{putri-etal-2024-llm}. 
Other work uses specialized translation models \citet{nllb2022} for quality check \citep{aakanksha-etal-2024-multilingual, nguyen-etal-2024-seallms}, or trains models to support low-resource dialects \citep{mousi-etal-2025-aradice}.\\
\textbf{LLM Generation} is increasingly used in dataset construction:
generating culturally specific content such as stereotypes \citep{jha-etal-2023-seegull, sahoo-etal-2024-indibias}, moral scenarios \citep{liu-etal-2024-evaluating-moral, dey-etal-2025-llms}, and norms \citep{ch-wang-etal-2023-sociocultural}; standardizing data formats \citep{chiu-etal-2025-culturalbench, umbet-etal-2025-kazbench}, and adapting content across cultural contexts \citep{joshi-etal-2025-cultureguard, putri-etal-2024-llm}. Beyond text, LLMs are used to synthesize images \citep{kim-etal-2025-tom}, generate image captions \citep{bai-etal-2025-power}, and label videos \citep{chen-etal-2025-videovista}.

\subsection{Insights into Dataset Creation Pipelines}

Our analysis reveals that modern cultural dataset creation rarely relies on a single, isolated method. Figure \ref{fig:datasets} illustrates common combinations of dataset creation approaches (additional details in Appendix \ref{sec:appendixdatapipeline}) with representative examples. We identify two dominant pipelines.\\
\textbf{Pipeline 1: Source-Grounded Human Curation} combines cultural knowledge sources with human expertise. 
Researchers typically adapt materials from books, image collections, and local websites, then ask humans to create prompts \citep{isbarov-etal-2025-tumlu}, questions \citep{nayak-etal-2024-benchmarking}, and QA pairs \citep{limkonchotiwat-etal-2025-wangchanthaiinstruct}, or annotate through quality checks \citep{kim-etal-2024-click, kim-lee-2025-nunchi}, label assignment \citep{maji-etal-2025-sanskriti, zhang-etal-2025-mllms}, or culture-specific translation \citep{winata-etal-2025-worldcuisines}, and enrich existing cultural datasets with expert knowledge \citep{cheng-etal-2025-hkcanto, romanyshyn-etal-2024-unlp}.\\
\textbf{Pipeline 2: LLM Co-Creation} uses LLMs as as generative \textbf{amplifiers} or \textbf{structuring tools} in dataset creation, with human input, theoretical frameworks, or existing language resources, including expanding seeded content with additional situated examples \citep{xu-etal-2025-self, zhan-etal-2024-renovi, li-etal-2025-multilingual}; generating culturally diverse content followed by human verification \citep{urailertprasert-etal-2024-sea, qiu-etal-2025-evaluating, ch-wang-etal-2023-sociocultural}; extending topic coverage \citep{wang-etal-2024-cdeval, cahyawijaya-etal-2025-high, chiu-etal-2025-culturalbench}; and extracting or generating additional information from existing language sources \citep{bhatia-etal-2025-swan, thorunn-arnardottir-etal-2025-wikiqa}.

Despite the common use of these pipelines, we find several exceptions across \textbf{Cultural Capabilities} (Figure \ref{fig:cc_by_data_creation_method}). Sociolinguistic Competence emphasize crowdsourced interactions to capture natural language use while avoiding explicit cultural cues that could create evaluation shortcuts.
Cultural Translation rely more on language professionals \cite{akinade-etal-2023-varepsilon, tapo-etal-2025-bayelemabaga, zhang-etal-2025-litransproqa}, and draw on web-based multilingual content, which captures contemporary expressions that often remain untranslated. 
Cultural Safety and Harm Reduction similarly depend on web data, particularly social media, to capture potentially harmful language needed for moderation \cite{maronikolakis-etal-2022-listening, mia-etal-2025-banmime}.
In contrast, Computational Cultural Representation use LLM-based synthetic generation to construct structured cultural knowledge that is difficult to obtain directly from raw data \cite{acquaye-etal-2024-susu, ziems-etal-2023-normbank, pujari-goldwasser-2025-llm}.

We also identify the \textbf{conceptual rigor trade-off} in the literature. Compared to other data construction approaches, culture research based methods are used less frequently. This pattern suggests that current cultural NLP dataset creation relies more heavily on coarse-grained, \textbf{proxy-based data} sources than on structured guidance from well-established \textbf{conceptual frameworks in social science}. Such a trend highlights a potential trade-off between scalability and practical convenience on the one hand, and depth, conceptual rigor, and precision of cultural representation on the other.

\input{figures/systems}

\section{Systems for Cultural Capabilities}
\label{sec:methods}

In this section we first discuss three different \textit{kinds} of systems that we identified during our survey (\S\ref{sec:methods_goals}), and then summarize some technical approaches adopted to improve CCs in NLP systems (\S\ref{sec:methods_technical}).

\subsection{Methodology Goals}
\label{sec:methods_goals}
Among all the papers that contributed a system, we identified three distinct overarching goals:\\
\textbf{System Improvement for CC:} Works that seek to directly \emph{improve} an NLP System's CCs, e.g., \cite{ma-etal-2025-enhancing, cao-etal-2025-specializing, li-etal-2024-schema}.\\
\textbf{System Evaluation for CC:} Works that propose a novel system to \emph{evaluate} an NLP System's CCs, e.g., \cite{zhao-etal-2025-makieval, mukherjee-etal-2023-global, zhang-etal-2025-cross}.\\
\textbf{Cultural Data Generation:} Works that propose a generalizable framework to \emph{generate} culturally informed data, e.g., \cite{keleg-magdy-2023-dlama, fung-etal-2023-normsage, hasan-etal-2025-nativqa}.

The three system goals each have a very different impact on research, which makes this an important distinction. We allow the ability to query papers by this distinction in \ui{}, and discuss insights from this classification in \S\ref{sec:insights}.

\subsection{Technical Approaches to Methodology}
\label{sec:methods_technical}
Fig.~\ref{fig:method_tree} illustrates the taxonomy used to classify methods based on their technical approaches. Prominent themes for each approach are discussed below.

\subsubsection{Prompting Based Methods}
We identified over 70 papers that contributed a novel system utilizing prompting in various ways.
One common approach is to vary the language of the prompt, to either probe a model \cite{kim-kim-2025-dual, zhao-etal-2025-makieval}, or to encourage culturally aligned behavior \cite{feng-etal-2024-teaching}.
Other methods use specialized prompts, such as injection of cultural knowledge \cite{shaikh-etal-2023-modeling, ma-etal-2025-enhancing}, guiding principles \cite{alkhamissi-etal-2024-investigating}, multi-step prompting \cite{hobson-etal-2024-story}, and cloze templates \cite{ramezani-xu-2023-knowledge}.

\subsubsection{Agent Systems: Retrieval and Multi-Model Systems}
We found Agent Systems, often with a "retriever agent" to be a common system proposed. What is the motivation for using an agentic framework? We find two recurring themes: (1) To allow for culturally diverse interactions and output \cite{ki-etal-2025-multiple, li-etal-2024-schema, white-etal-2024-communicate, bai-etal-2025-power, feng-etal-2024-modular}, (2) To allow for specialized roles in complex systems made up of smaller modules \cite{anik-etal-2025-preserving, wu-etal-2024-transagents, yuan-etal-2024-measuring, liang-etal-2025-colloquial}. For systems that utilized a retriever, what was the role of the retriever? The most common role we found was to retrieve diverse kinds of cultural information such as entities \cite{conia-etal-2024-towards}, recipes \cite{hu-etal-2024-bridging}, medical text \cite{calvo-bartolome-etal-2025-discrepancy}, poetry \cite{chen-etal-2025-benchmarking-llms}. This would be explained by the fact that the most common CC associated with retrievers was Cultural Translation.

\subsubsection{Human in the Loop Systems}
We identified 13 papers that proposed a system involving a human in the loop. The main role of humans in these systems was to provide cultural expertise, which can look like judgments and corrections \cite{pujari-goldwasser-2025-llm, ziems-etal-2025-culture}, culturally informed seed data \cite{hasan-etal-2025-nativqa, rachamalla-etal-2025-pragyaan, putri-etal-2024-llm}, adversarial input \cite{chiu-etal-2025-culturalbench}, and even outside culture judgment \cite{park-etal-2025-llm}.

\subsubsection{Training Based Methods}
Many papers propose novel training paradigms to improve the CC of NLP Systems. By far the most common one we identify is SFT on LLMs for some form of cultural alignment \cite{choenni-etal-2024-echoes, cao-etal-2025-specializing, xu-etal-2025-self, dai-etal-2025-word}. Another common trend we notice is that of training smaller models for specialized tasks, such as Bert \cite{devlin2019bertpretrainingdeepbidirectional} to identify values \cite{kiesel-etal-2022-identifying} or predict stereotypes \cite{kim-johnson-2025-korean}, e5 model \cite{wang2024multilinguale5textembeddings} to align sociocultural concepts \cite{wang-etal-2025-dlir}, or evaluation metrics trained to reflect human judgment \cite{bayramli-etal-2025-diffusion}. Finally, there are several papers that introduce novel architectures for specialized tasks such as subjective prediction \cite{parappan-henao-2025-learning}, predicting social relationships from videos \cite{zhang-etal-2025-interesting}, and transfer learning across languages \cite{ringel-etal-2019-cross}.

\subsubsection{Classical NLP}
In this subsection we discuss approaches which are often paired; embeddings, lexica, and methods over them like clustering, summarization.\\
\textbf{Lexica:} We identified 2 main ways that lexica are used in this field: (1) discovering cultural differences like ideologies \cite{milbauer-etal-2021-aligning}, perspectives \cite{gutierrez-etal-2016-detecting}, expressions of mental health \cite{rai-etal-2025-cross}, and personal values \cite{wilson-etal-2016-disentangling}, and (2) quantifying linguistic aspects like style \cite{havaldar-etal-2023-comparing}, bias, \cite{naous-xu-2025-origin, friedman-etal-2019-relating} cultural awareness \cite{zhao-etal-2025-makieval, caplan_splits_2025}, harm \cite{menis_mastromichalakis_dont_2025}, and concepts \cite{li-zhang-2023-cultural}.\\
\textbf{Embeddings:} Lexical methods are often paired with embedding based methods for purposes like enriching the lexica \cite{havaldar-etal-2024-building, havaldar-etal-2023-comparing}, comparing similarities or differences between cultures \cite{sun-etal-2021-cross, milbauer-etal-2021-aligning}, and detecting cultural biases \cite{friedman-etal-2019-relating}. In contrast, \cite{rai-etal-2025-cross, caplan_splits_2025} specifically avoid embeddings to prevent issues arising from bias in embeddings. Other works, exploit the bias in embeddings to measure abstract concepts like values, ideologies and identities \cite{cahyawijaya-etal-2025-high, milbauer-etal-2021-aligning, ventura-etal-2025-navigating, havaldar-etal-2024-building}, and perceptions and pragmatics \cite{sun-etal-2021-cross, lin-etal-2018-mining, white-etal-2024-communicate}. Another common use of embeddings is to create a shared representation space for multiple cultures, in order to discover similarities and differences between them \cite{choenni-etal-2024-echoes,zhou-etal-2023-cultural}.\\
\textbf{Statistical Methods:} These methods encompass a wide variety of tools like clustering, topic modeling, evaluation metrics etc. We highlight two interesting recurring themes; the first is the use of clustering and topic modeling to discover common themes from large noisy data \cite{cuevas-etal-2025-anecdoctoring, milbauer-etal-2021-aligning, hobson-etal-2024-story, pujari-goldwasser-2025-llm}, and the second is the use of various metrics to measure abstract concepts like cultural alignment \cite{wang-etal-2024-countries}, ideologies \cite{milbauer-etal-2021-aligning}, and values \cite{xu-etal-2024-exploring-multilingual}.

\subsubsection{Mechanistic Interpretability}
We identified few methods using mechanistic interpretability, which indicates it being an underexplored method for cultural NLP. The papers we did identify propose novel ways to answer why models display the cultural tendencies that they do \cite{xu-etal-2024-exploring-multilingual, ying-etal-2025-disentangling}.

\subsubsection{Multimodal Methods}
Several papers we surveyed proposed methods that included modalities beyond text. A very interesting line of work uses vision to account for low resource languages that do not have a surplus of textual data \cite{li-zhang-2023-cultural, chen-etal-2024-logogramnlp, r-etal-2025-field}. Another motivation for such methods is cultural competence over non-text modalities like vision \cite{sharma-etal-2022-disarm, alwajih-etal-2024-peacock, zhang-etal-2025-llm-driven, zhang-etal-2025-interesting, khanuja-etal-2025-towards}.

\section{Observations and Recommendations for Future Work}
\label{sec:insights}

\subsection{Trends in Proposed Future Work}

We analyzed the future work sections of the surveyed papers and found that community-proposed directions predominantly fall into two themes (details in Appendix \ref{app:futurework}). First, researchers frequently call for \textbf{scaling cultural and linguistic coverage} to address data scarcity, often advocating for richer, multimodal benchmarks \cite{wang-etal-2024-seaeval, maji-etal-2025-sanskriti}. Second, there is a strong push to \textbf{model cultural complexity} more accurately. Authors emphasize moving away from static, monolithic labels by operationalizing culture as a dynamic distribution \cite{havaldar-etal-2023-comparing, ziems-etal-2023-normbank}, incorporating intersectional and fine-grained geographic variables beyond language \cite{falk-etal-2024-overview, koto-etal-2024-indoculture}, and developing evaluation frameworks explicitly accounting for pluralism \cite{zhou-etal-2023-cultural, miehling-etal-2025-evaluating}.

\subsection{Our Observations and Recommendations}
\label{sec:our-observations}

In this section we discuss four broader patterns that offer promising avenues for the community.

\paragraph{Prioritizing system improvements over additional dataset creation.}
Dataset creation substantially outpaces the development of methods that directly improve CCs. This gap is especially pronounced for Cultural Knowledge (>100 surveyed \textbf{dataset} contributions), Cultural Safety and Harm Reduction (>60), and Value-driven Cultural Alignment (>50), each of which has fewer than 20 surveyed \textbf{system} improvement contributions. The abundance of such datasets \cite{singh-etal-2025-global, sahoo-etal-2025-diwali, bui-etal-2025-multi3hate, shetty-etal-2025-vital} indicates that the community has built a robust foundation for measuring these capabilities. We recommend that future efforts emphasize developing systems specifically designed to improve performance on these existing datasets, as done by \cite{ki-etal-2025-multiple, wang-etal-2025-multilingual, feng-etal-2024-teaching, parappan-henao-2025-learning}.

\paragraph{Reusing existing data generation frameworks for scalable data collection.}
When the creation of new data is desired, we observe an opportunity to reduce redundant effort. We identified several sophisticated, generalizable Cultural Data Generation systems \cite{ziems-etal-2025-culture, caplan_splits_2025}. These frameworks are designed to produce datasets dynamically by conditioning on target variables (e.g., country, language, or source collection). However, we noticed limited follow-up reuse of these pipelines. Leveraging these existing generative frameworks can accelerate research when specific cultural data is scarce, providing a scalable alternative to curating datasets from scratch.

\paragraph{Developing specialized evaluation methods for implicit cultural nuances.}
Our survey suggests that measurement paradigms for different CCs are evolving at different rates. We use the ratio of novel \textit{Evaluation System} papers to \textit{Dataset} papers to approximate this focus: a higher ratio indicates active development of bespoke measurement frameworks, whereas a lower ratio implies a reliance on existing metrics. For example, capabilities involving abstract constructs, such as Value-driven Cultural Alignment (0.30) and Sociolinguistic Competence (0.21), exhibit relatively high ratios. This suggests that \textbf{standard metrics often fall short} for these CCs, prompting researchers to build specialized evaluators \cite{yao-etal-2024-value, shen-etal-2025-mind, casola-etal-2024-multipico, ying-etal-2025-disentangling}. Conversely, capabilities like Multimodal Cultural Knowledge (0.02) and Cultural Translation (0.11) show significantly lower ratios, as they frequently \textbf{rely on established, reference-based metrics} like BLEU or ROUGE. While these standard metrics provide valuable baselines, we encourage the continued development of specialized evaluation methodologies for these domains as well.

This opportunity is particularly visible in multimodal work. Existing literature provides excellent coverage of visually salient cultural artifacts, such as food, clothing, and landmarks \cite{nayak-etal-2024-benchmarking, bhatia-etal-2024-local, maji_drishtikon_2025}. Moving forward, the field is well-positioned to tackle subtler, ``below-the-iceberg'' phenomena \cite{hall1976beyond} essential for real-world interaction, such as conversational grounding, gestures, paralinguistic interpretation, and cross-dialect understanding, e.g., \cite{sasu-etal-2025-akan, zhang-etal-2025-interesting}.

\paragraph{Evaluating and optimizing multiple CCs jointly.}
Our taxonomy defines CC as a collection of distinct competencies (\S\ref{subsec:cc-tax}), yet current research predominantly isolates them. Among surveyed papers contributing \textit{system improvements}, 82\% target exactly one CC; for \textit{system evaluation}, the figure is 94\%. While this focused scope is necessary for foundational research, real-world deployments require systems to juggle \textbf{multiple CCs simultaneously}. Improvement on one CC does not inherently transfer to another \cite{zhang_culturescope_2025}; a culturally capable system deployed in the world may need to retrieve accurate cultural facts, communicate appropriately, and adapt to pluralistic norms concurrently. We encourage future work to report cross-CC transfer and trade-offs, and to develop comprehensive benchmarks that evaluate multiple CCs jointly.

\section{\ui{}: Community Resource}
\label{sec:culturemine}
In this section we provide a brief overview of \ui{}, which contains all the papers we surveyed, tagged with the dataset taxonomy and/or the systems taxonomy and relevant CCs. Additionally, each paper is tagged for culturally relevant metadata such as what proxy it used for culture (language, country, other). When applicable, papers are also tagged for the exact languages or regions that they account for. All the papers can be queried via drop-down filters, which makes it very easy for researchers to find a collection of work that is relevant for them! The top section of \ui{} has a dynamic visualization giving an overview of papers and updates based on selected fields.

\begin{figure}[t]
    \centering
    \includegraphics[width=\columnwidth]{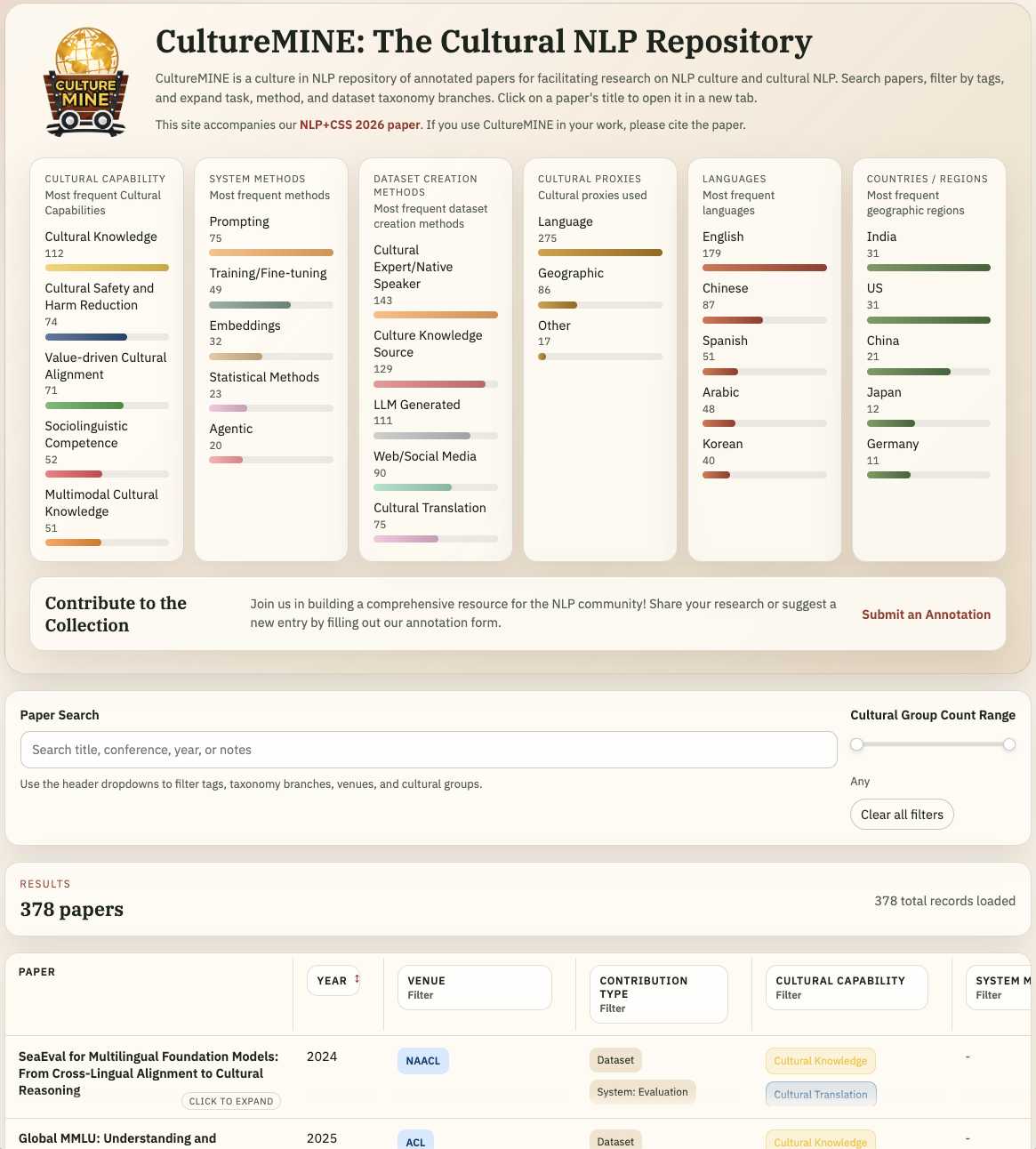}
    \caption{\ui{} interface has dynamic visualizations in the top section. The fields in the bottom, which include our taxonomies for datasets and systems, can be used as filters to find relevant papers. The filters also include information such as cultural proxies used, what cultures, and what languages!}
    \label{fig:culture-mine}
\end{figure}

Additionally, we include a submission form that researchers can fill out to add more papers. We intend to keep the UI updated regularly, and hope that it can be a true mine for cultural NLP!

\section{Conclusion}
We surveyed over 375 papers, provided an overview of recent technical methods for improving the CCs of NLP systems, proposed taxonomies to organize contributions and released our surveyed papers, tagged with our taxonomies and other rich metadata via \ui{}. We hope this survey and \ui{} will be a valuable resource for the cultural NLP community.

\section*{Limitations}

Despite our effort to provide a comprehensive overview of the cultural capabilities of NLP systems, several limitations remain.

\paragraph{Scope and Source Coverage.} 
Our survey is not exhaustive. Research on culture in NLP is distributed across multiple venues and related fields, making comprehensive coverage infeasible. We therefore focus on papers that directly address our central question: how cultural capabilities are technically represented, operationalized, or incorporated in NLP systems and research designs. For the surveyed papers, we systematically collected them from the ACL Anthology, which serves as the primary repository. As a result, research published outside the ACL Anthology may be underrepresented in our survey.

\paragraph{Exclusion of NLP for Social Science.} 
An important related area not included in this survey is NLP for social science. This line of research applies natural language processing methods to study social phenomena such as cultural variation, social meaning, and contextual interpretation in large-scale text data. Such work provides valuable theoretical and empirical perspectives for computational research on culture. Establishing a more systematic connection between the NLP for social science literature and culture-focused NLP research remains an important direction for future work.

\paragraph{Taxonomic Boundaries.} 
The taxonomy proposed in this survey is derived through a bottom-up analysis of the papers we reviewed. It is not intended to be a complete theory of all possible mechanisms for investigating cultural capabilities in NLP. As the literature develops, additional dimensions and categories may emerge.

\paragraph{Interpretative Constraints.} 
Our analysis is constrained by how culture is defined and described in the surveyed papers. Cultural modeling choices are often only partially formalized or embedded within broader task or dataset design decisions. Consequently, some categorization decisions require interpretation.

\paragraph{Conceptual Proxy Limitations.} 
Many studies represent culture through proxies such as language, geographic region, nationality, or demographic group. These proxies capture different aspects of culture and are not conceptually equivalent. Our survey organizes the current technical design space, but it does not resolve the underlying conceptual challenges of modeling culture.

\section*{Ethical Considerations}

Culture is complex, dynamic, and internally heterogeneous. However, computational studies often operationalize culture using simplified proxies such as language, nationality, geographic region, or demographic group. These proxies can be practically useful for empirical analysis, but they may also reify culture as fixed or homogeneous, obscure within-group diversity, and inadvertently reinforce stereotypes or exclusions.

The goal of this survey is to organize and analyze the technical mechanisms through which prior work incorporates culture into NLP systems. We do not treat any particular operationalization of culture as definitive or exhaustive. Rather, we view these operationalizations as modeling choices made for specific empirical purposes. We therefore encourage future research to make these assumptions explicit, carefully justify the cultural proxies used, and evaluate the potential downstream risks associated with cultural generalization in NLP systems.

\section*{Acknowledgments}
We thank the anonymous reviewers and Rajkumar Pujari, for their insightful comments that helped improve the paper. This project was partially funded by NSF IIS-2048001.

\bibliography{custom}

\appendix

\begin{figure}[htb]
    \includegraphics[width=\linewidth]{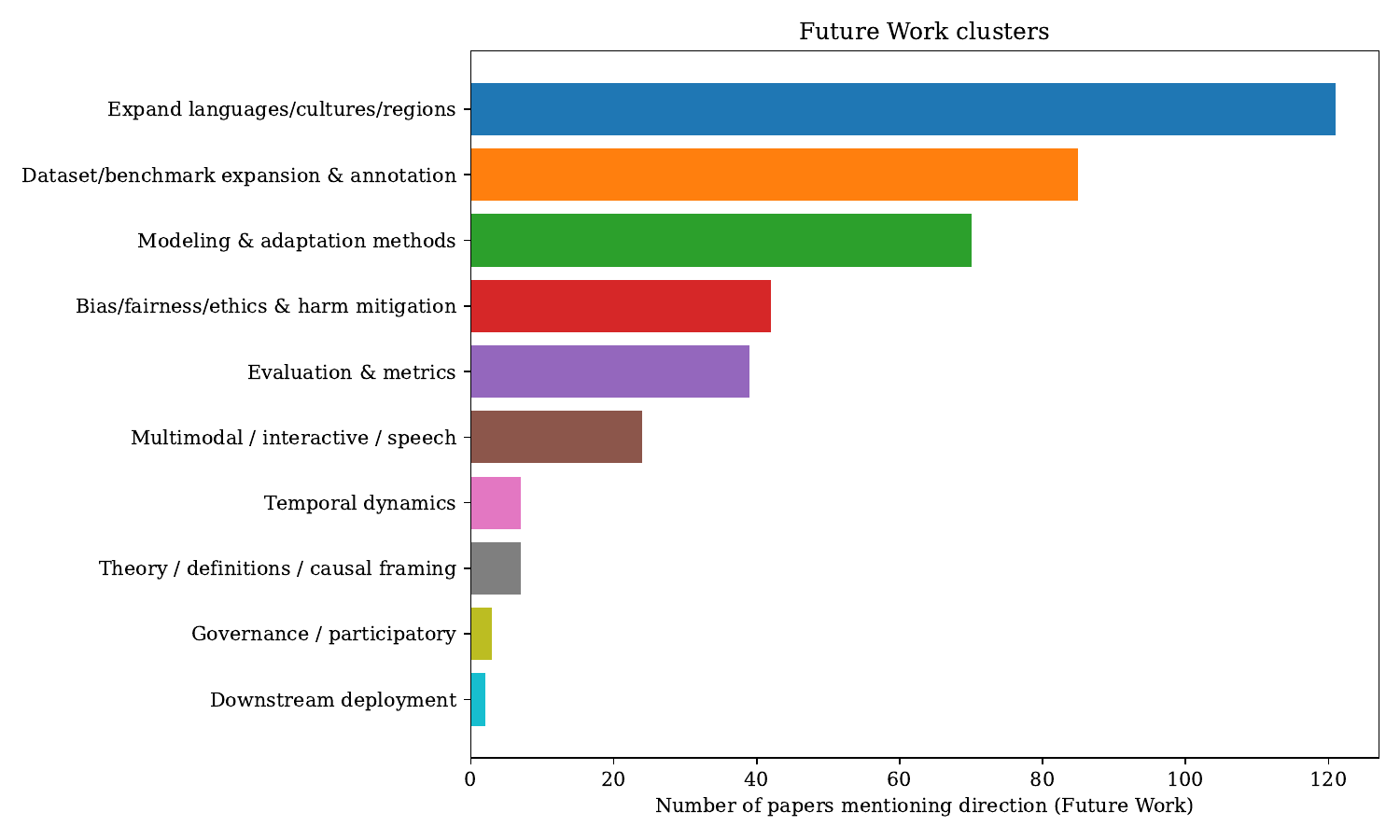}
    \caption{Frequency of future-work clusters in the papers}
    \label{fig:future_work}
\end{figure}

\section{Detailed Literature Collection and Annotation Methodology}
\label{app:collection}
\subsection{Phase 1: Initial Seed Collection}
We took a two-phased approach to collect relevant literature, focusing on the past five years (including all available papers from 2025) to capture the modern landscape of Cultural NLP. The first phase consisted of a manual search combined with an automated scrape of the ACL Anthology. We queried for papers containing the words ``culture'' or ``cultural'' in the title, alongside at least one other keyword related to investigating or operationalizing culture in NLP. This initial highly targeted search resulted in a seed set of 180 papers.

\subsection{Phase 2: Broad Scrape and LLM-Assisted Filtering}
To ensure comprehensive coverage, the second phase involved a broader scrape of the ACL Anthology for any paper containing ``culture'' or ``cultural'' in either the title or abstract, returning 1,216 additional candidates. Given the volume, we employed an LLM to assist in filtering the list down to a highly relevant subset.

First, we prompted the LLM with a description of our inclusion criteria along with the title and abstract of each paper, retaining only those assigned a high relevance score. Next, we prompted the LLM with stricter inclusion guidelines, providing it with potentially relevant full-text sections of each candidate paper (e.g., Introduction, Dataset, Methods, Evaluation). To ensure we analyzed works with thorough descriptions and evaluations of proposed datasets or methods, we filtered out short papers, considering only those longer than 6 pages. This rigorous second phase yielded 277 additional highly relevant papers.

\subsection{Manual Annotation and Consensus Strategy}
The combined pool of candidate papers was manually processed by annotators who are Computer Science PhD students actively involved in NLP research. Papers were annotated across three main dimensions: (1) Cultural Capabilities addressed, (2) dataset creation methodology, and (3) technical system methods utilized.
Throughout the annotation process, we performed a manual filtering step to discard papers that did not pose a novel contribution in any of these three aspects. Specifically, we removed works that solely evaluated existing models on existing datasets, or papers utilizing NLP tools for pure sociolinguistic or social science analyses without proposing a technical system or dataset improvement. This exclusion phase removed 79 papers, leaving a final corpus of 378 papers to be read thoroughly and annotated.

To ensure high-quality and consistent annotations, regular meetings were held between the surveyors. During these meetings, precise definitions of the labels were discussed, disagreements were resolved, and edge cases were evaluated. We also conducted cross-validations on small subsamples of papers to purposefully identify and resolve points of contention. Once this iterative process concluded and the taxonomy definitions were firmly established, the team went back and corrected all previously labeled papers to guarantee strict consistency and adherence to the final taxonomy across the entire 378-paper corpus.

\section{Future Work Cluster Analysis}

\label{app:futurework}

We analyze the Future Work sections recorded in our annotated spreadsheet. The aggregated results are summarized in Figure~\ref{fig:future_work}. We identify recurring directions through a transparent and reproducible clustering procedure. Because individual papers often propose multiple extensions, clustering is treated as a multi-label assignment problem. A single future work entry can therefore belong to multiple clusters.

\begin{figure}[]
    \includegraphics[width=0.9\linewidth]{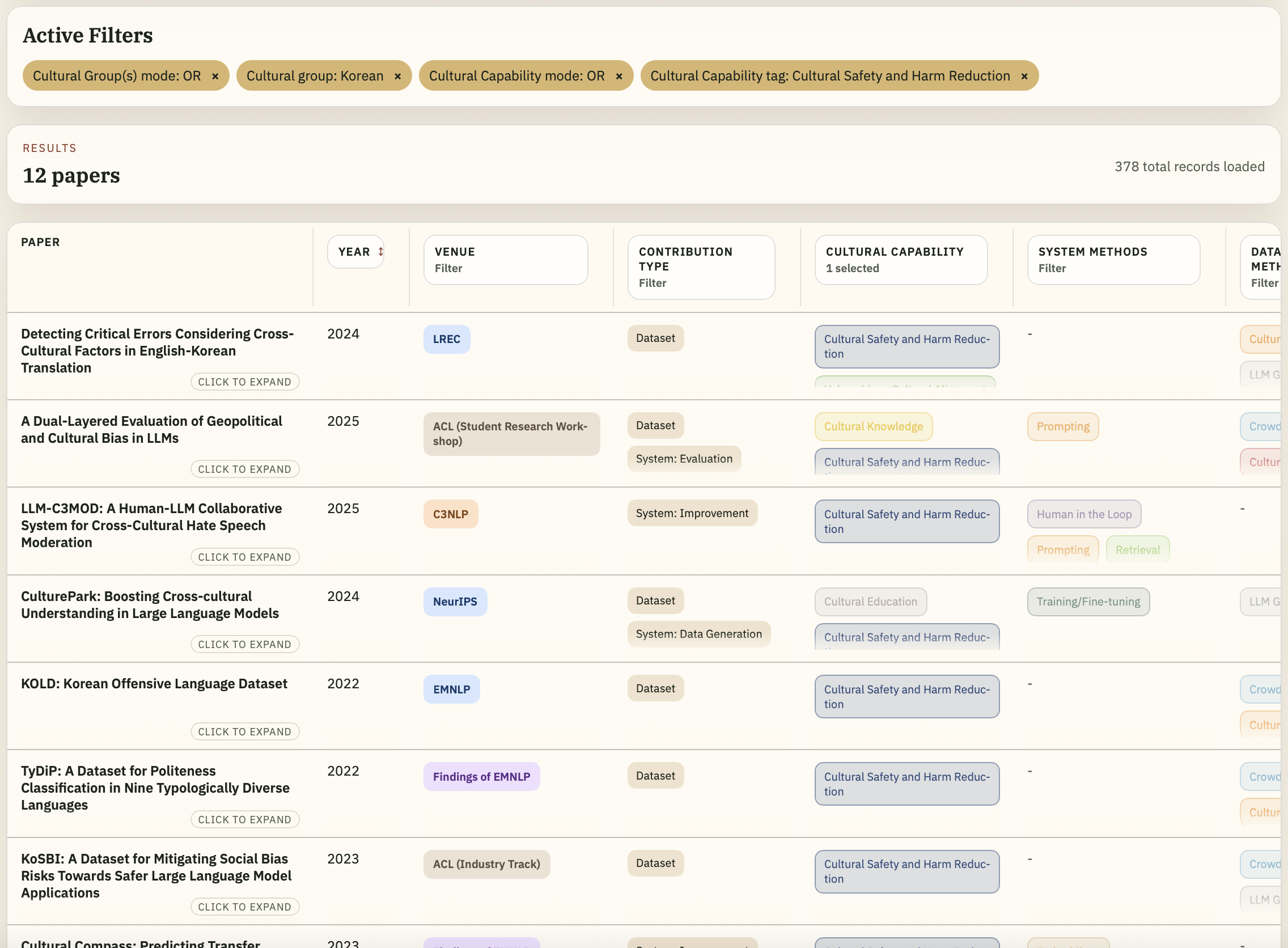}
    \caption{Filtered Paper List in \ui}
    \label{fig:ui_filtered_papers}
\end{figure}

\subsection{Cluster schema}
We define a small set of coarse themes that repeatedly appear across the surveyed corpus. Each cluster represents a practical research direction rather than a topical category. The clusters and their descriptions are as follows:
\begin{itemize}
    \item \textbf{Expand languages and cultures}: extending coverage to additional languages, dialects, regions, or cultural contexts.
    \item \textbf{Dataset and benchmark expansion}: collecting new datasets, enlarging benchmarks, adding annotations, or improving dataset curation and documentation.
    \item \textbf{Modeling and adaptation methods}: developing improved modeling, training, fine-tuning, alignment, transfer, retrieval, or adaptation approaches that explicitly address cultural phenomena.
    \item \textbf{Bias, fairness, ethics, and harm mitigation}: measuring or mitigating culturally shaped bias, stereotyping, toxicity, or other harmful outputs.
    \item \textbf{Evaluation and metrics}: proposing stronger evaluation metrics, human evaluation protocols, robustness checks, or generalization tests for cultural capabilities.
    \item \textbf{Multimodal or interactive settings}: extending cultural modeling to speech, audio, vision, dialogue systems, or other interactive environments.
    \item \textbf{Temporal dynamics}: modeling culture as a time-varying process, including updates, drift, or longitudinal change.
    \item \textbf{Theory, definitions, and causal framing}: clarifying conceptual definitions of culture or proposing causal or mechanistic interpretations of cultural variables.
    \item \textbf{Governance and participatory approaches}: incorporating community participation, consent mechanisms, stewardship practices, or governance frameworks.
    \item \textbf{Downstream deployment}: evaluating integration into real-world systems and conducting deployment-oriented validation.
\end{itemize}

\subsection{Assignment procedure}

Cluster assignments are produced using a dictionary-based matching procedure. For each cluster, we construct a small set of indicative surface forms, such as ``multilingual'', ``dialect'', ``benchmark'', ``annotation'', ``robustness'', ``toxicity'', ``speech'', or ``drift''. A cluster is marked as present when any of its indicative patterns appear in the corresponding future work text using case-insensitive matching. This design prioritizes interpretability and reproducibility: cluster definitions are explicit, and assignments can be verified by inspecting the matched expressions.

\begin{figure}[]
    \includegraphics[width=\linewidth]{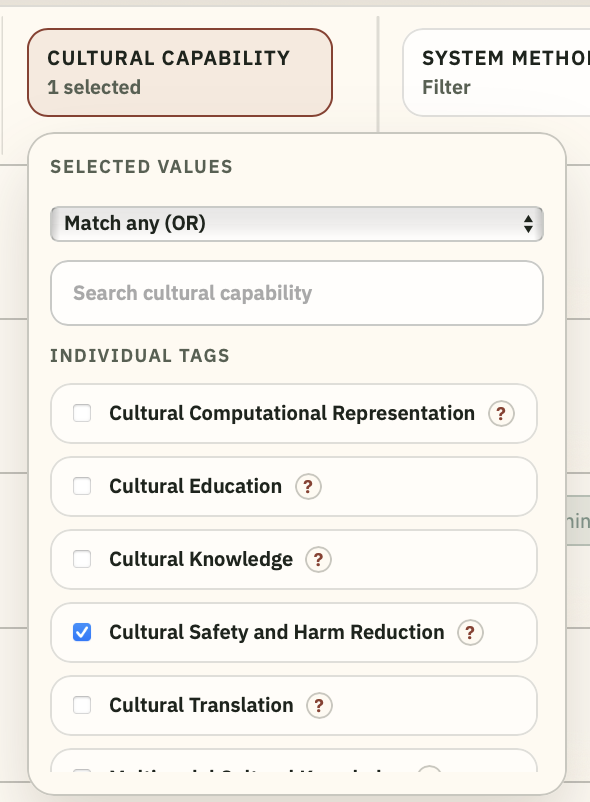}
    \caption{Cultural Capabilities Filter in \ui}
    \label{fig:ui_cc_filter}
\end{figure}

The clustering procedure is intentionally lightweight and surface form-driven. It may undercount papers that describe a direction using uncommon phrasing, and it does not disambiguate cases where a keyword appears with a different meaning. We therefore treat the clusters as a descriptive summary of recurring future work themes rather than as an exhaustive taxonomy.

\section{Analysis of Methods used for CCs}
\subsection{Methods across cultural capabilities}

\begin{figure*}[ht]
    \centering
    \includegraphics[width=\linewidth]{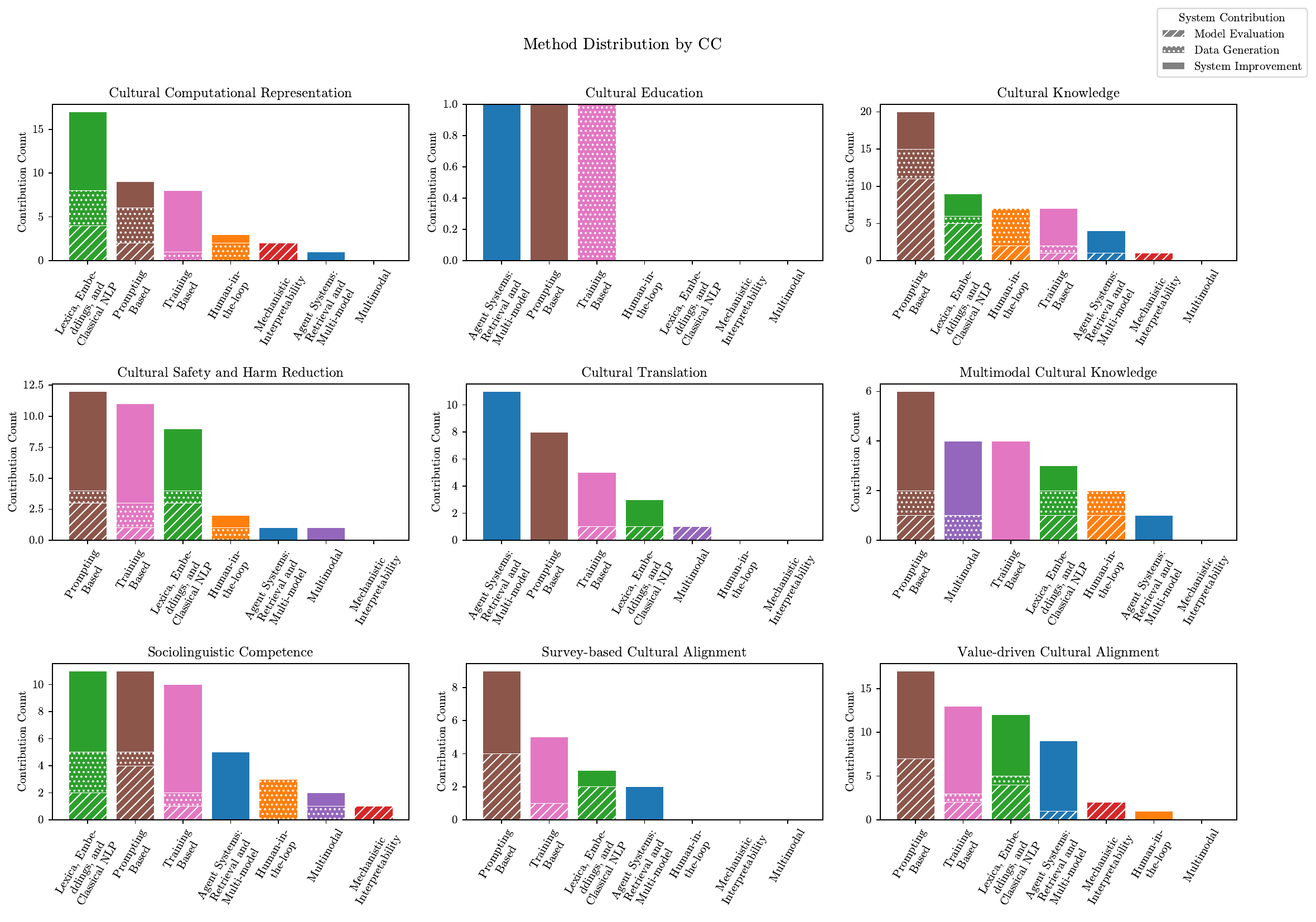}
    \caption{Method distribution across cultural capabilities}
    \label{fig:methods_by_cc}
\end{figure*}

Figure~\ref{fig:methods_by_cc} presents a $3\times3$ grid of stacked bar charts comparing method families across cultural capabilities. Each subplot corresponds to one cultural capability. The x-axis lists methodological families, while the y-axis reports the number of contributions. Each bar is partitioned into segments representing four contribution types: System Contribution, Model Evaluation, Data Generation, and System Improvement. Overall, Cultural Knowledge and Value-driven Cultural Alignment contain the largest number of contributions, whereas Cultural Education, Survey-based Cultural Alignment, and Multimodal Cultural Knowledge appear relatively sparse. Across cultural capabilities, prompting-based methods are the most frequently used approach, with training-based methods and lexica, embeddings, or other classical NLP techniques also appearing regularly. Some cultural capabilities display clearer methodological specialization. Cultural Computational Representation relies more heavily on classical NLP-style approaches, while Cultural Translation more often adopts agent-based and prompting-based strategies.

\begin{figure}[]
    \includegraphics[width=\linewidth]{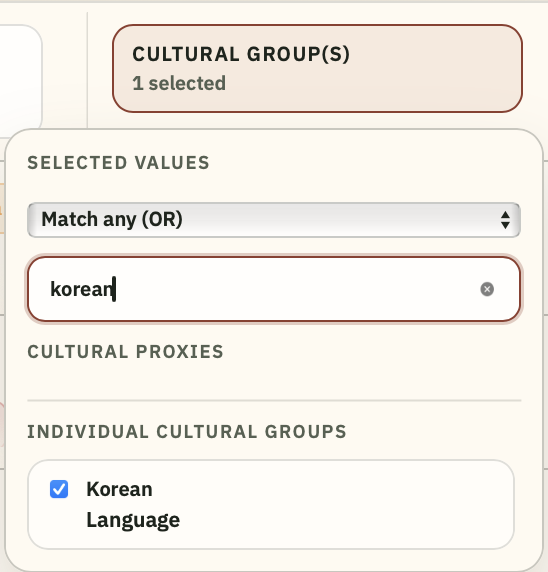}
    \caption{Cultural Groups Filter in \ui}
    \label{fig:ui_culture_group_filter}
\end{figure}

A key observation is that cultural NLP research is methodologically heterogeneous. Different tasks tend to attract different technical approaches rather than converging on a single dominant paradigm. Current work is largely centered on prompting and training-based techniques, particularly for tasks involving cultural knowledge, value alignment, and safety, reflecting the broader influence of large language models in recent NLP research. In contrast, tasks related to cultural representation and sociolinguistic competence maintain stronger connections to earlier NLP traditions. The figure also highlights several relatively underexplored areas, including cultural education, multimodal cultural reasoning, and mechanistic interpretability, indicating opportunities for future work to expand both task coverage and methodological diversity.

\subsection{Cultural capabilities across methods}

\begin{figure*}[ht]
    \centering
    \includegraphics[width=\linewidth]{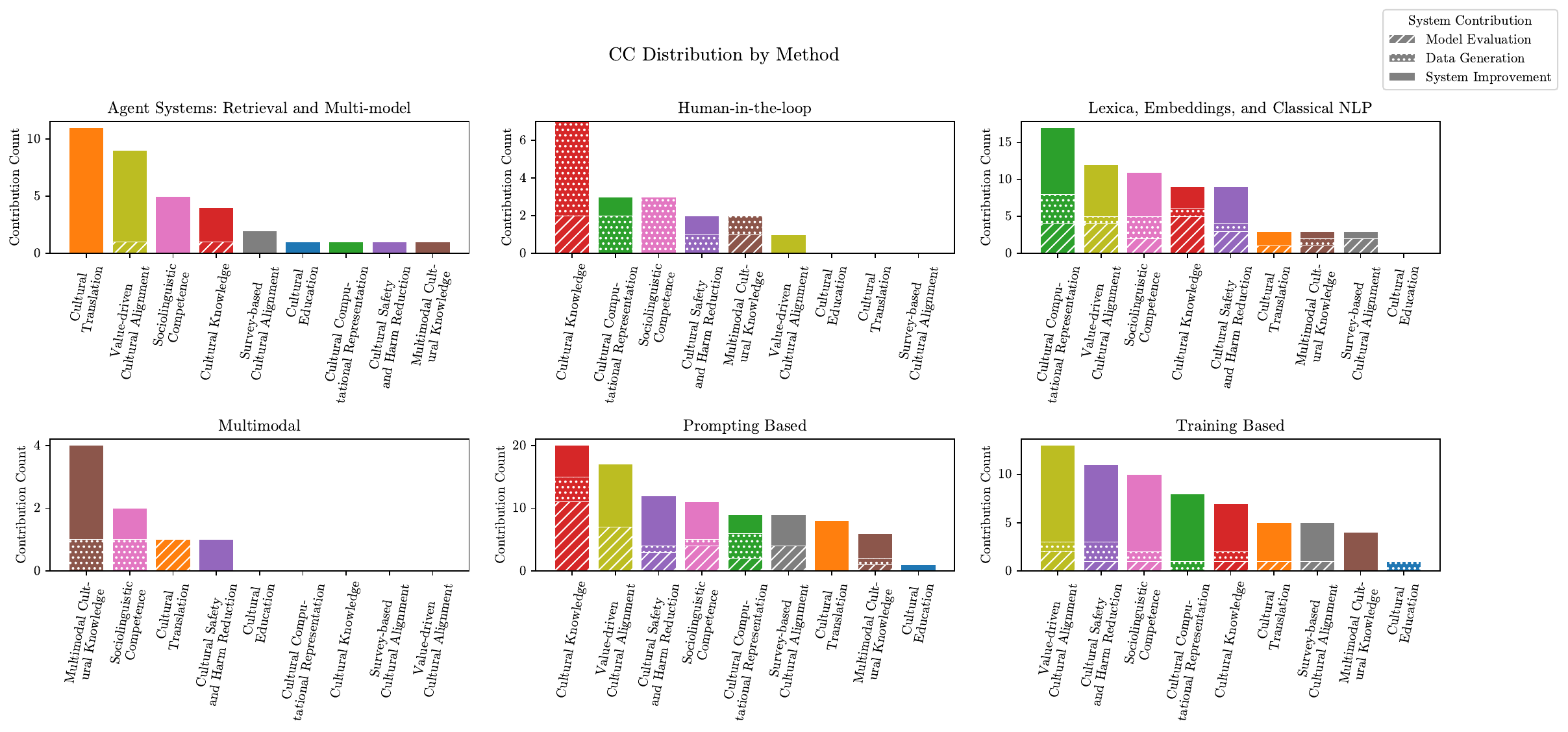}
    \caption{Cultural capability distribution across methods}
    \label{fig:cc_by_methods}
\end{figure*}

Figure~\ref{fig:cc_by_methods} presents the distribution of cultural capabilities across method families using a $2 \times 3$ grid of bar charts. Each subplot corresponds to a method family, and the horizontal axis lists cultural capabilities, while the vertical axis reports the number of contributions. The six method families are retrieval and multimodel approaches, human-in-the-loop methods, lexica, embeddings, classical NLP techniques, multimodal approaches, prompting-based methods, and training-based methods. Across methods, prompting-based approaches exhibit the largest overall volume and cover the widest range of cultural capabilities, followed by training-based methods and lexica, embeddings, and classical NLP techniques. In contrast, human-in-the-loop and multimodal approaches account for substantially fewer contributions and are concentrated in a limited set of cultural capabilities. At the capability level, cultural knowledge, value-driven cultural alignment, sociolinguistic competence, and cultural computational representation appear repeatedly across multiple method families. Cultural education and survey-based cultural alignment remain comparatively limited. The stacked bar segments indicate that many methods support multiple contribution types, including system contribution, model evaluation, data generation, and system improvement, rather than focusing on a single contribution category.

\begin{figure*}[ht]
    \centering
    \includegraphics[width=\linewidth]{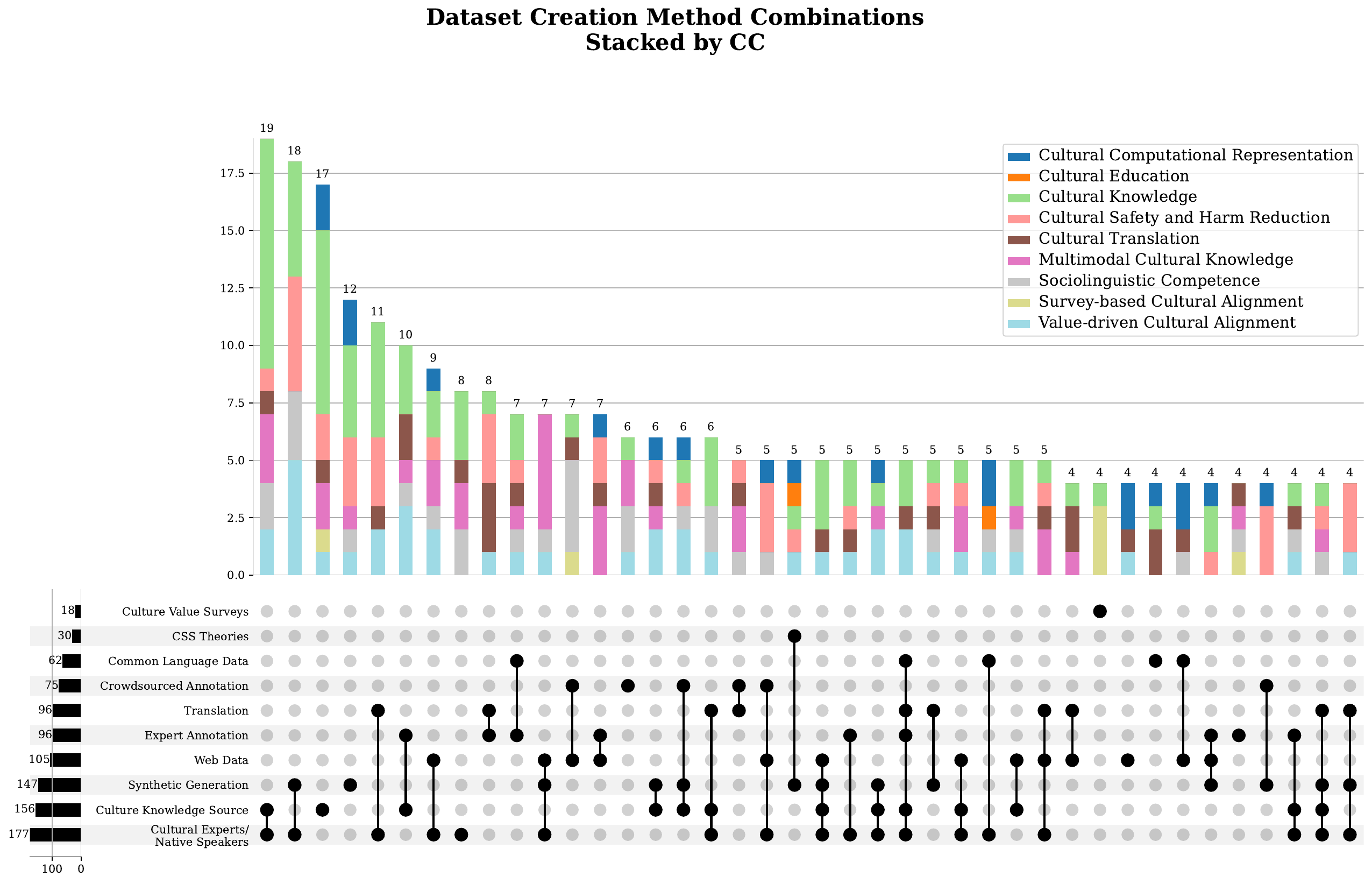}
    \caption{Dataset creation method combinations stacked by cultural capability}
    \label{fig:data_by_cc}
\end{figure*}

The figure further suggests that the field is structured around a small set of widely reusable technical paradigms, particularly prompting-based and training-based approaches, which support a broad range of cultural capabilities. This pattern indicates that much of cultural NLP research adapts general-purpose large language model techniques to culture-related problems rather than developing highly task-specific architectures. At the same time, method families differ in flexibility. Lexica, embeddings, and classical NLP techniques remain particularly relevant for representational and sociolinguistic analysis, whereas multimodal and human-in-the-loop approaches appear more specialized and comparatively underexplored. Overall, the distribution indicates that research diversity varies not only across cultural capabilities but also across technical methods, with several approaches functioning as central methodological hubs while others remain peripheral.

\subsection{Cultural Capabilities Across Dataset Creation Methods} \label{sec:appendixdatapipeline}
\begin{figure}[ht]
    \centering
    \includegraphics[width=\linewidth]{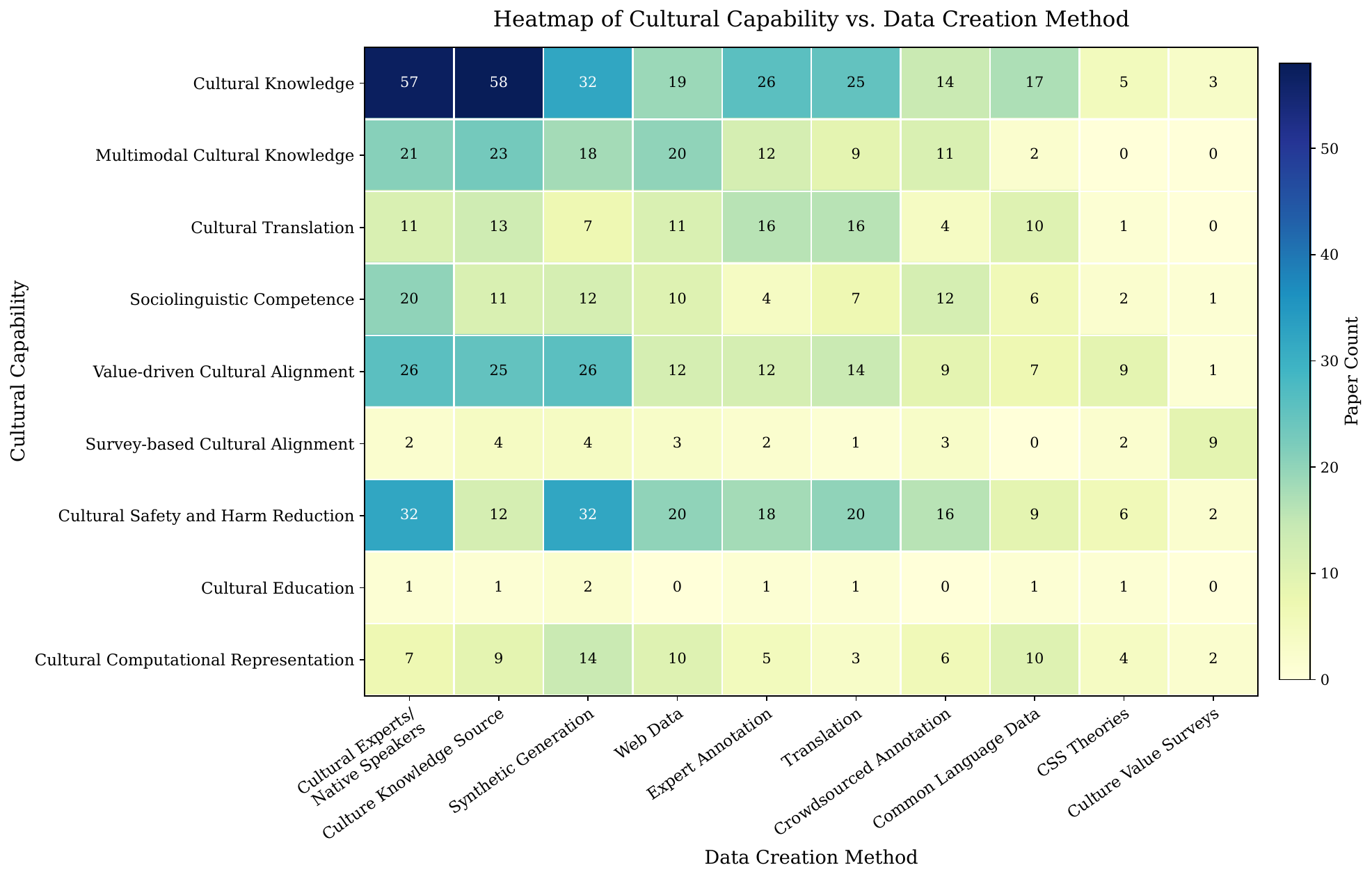}
    \caption{Heatmap of cultural capability by dataset creation method}
    \label{fig:cc_by_data_creation_method}
\end{figure}

Figure~\ref{fig:data_by_cc} presents an UpSet-style visualization of dataset creation method combinations, stacked by cultural capability. The bottom matrix indicates which dataset creation ingredients appear in each combination pattern. These ingredients include Cultural Experts or Native Speakers, Culture Knowledge Source, Synthetic Generation, Web Data, Expert Annotation, Translation, Crowdsourced Annotation, Common Language Data, Cultural and Social Science Theories, and Culture Value Surveys. The bars above the matrix report the frequency of each ingredient combination, with segments colored by cultural capability category. The bars on the left indicate the marginal frequency of individual ingredients. Cultural Experts or Native Speakers, Culture Knowledge Source, Synthetic Generation, and Web Data appear most frequently, while Cultural and Social Science Theories and Culture Value Surveys occur relatively rarely. At the combination level, the most frequent patterns reach approximately the high teens, while many other combinations appear only four to six times. Across these combinations, Cultural Knowledge contributes the largest share in many of the most common patterns, while Value-driven Cultural Alignment, Sociolinguistic Competence, and Cultural Translation also appear across multiple combinations.

\begin{figure}[]
    \includegraphics[width=\linewidth]{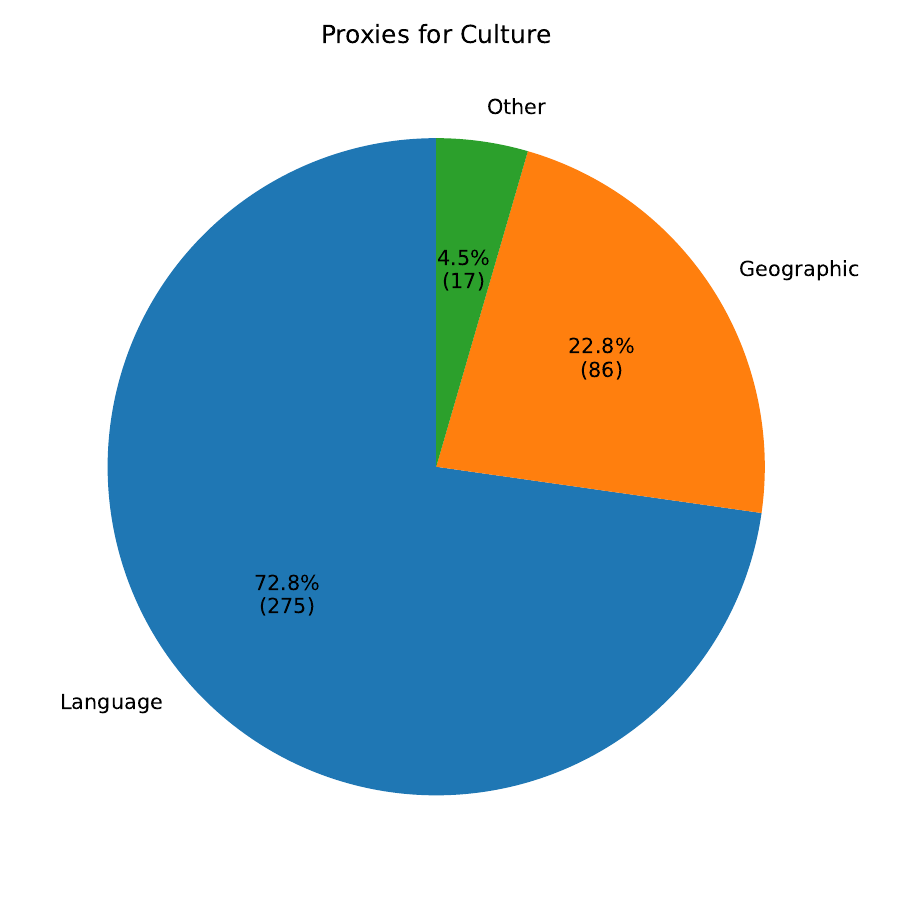}
    \caption{Pie Chart of Cultural Proxies Used}
    \label{fig:proxy_pie}
\end{figure}

Figure~\ref{fig:cc_by_data_creation_method} shows a heap map of dataset creation method vs CCs. A key observation is that dataset creation for cultural natural language processing is typically compositional rather than based on a single source. Many studies construct datasets by combining several ingredients, most commonly human expertise from native speakers, external cultural knowledge resources, synthetic data generation, and web-collected data. This indicates that cultural information is rarely operationalized through a single data pipeline. Instead, researchers assemble datasets by integrating human knowledge, structured cultural resources, and scalable generation strategies. The figure also reveals an imbalance in the maturity of different cultural data construction approaches. High-frequency combinations are dominated by Cultural Knowledge-oriented pipelines, suggesting that dataset development has progressed most strongly for knowledge-centered tasks. In contrast, theoretically grounded sources such as cultural value surveys and cultural or social science theories remain uncommon. Overall, the distribution of combinations suggests that current practice prioritizes pragmatic and scalable data construction strategies, while theoretically grounded but less reusable approaches remain underexplored.

\section{\ui}
\label{app:ui}

We accompany our survey with a web-interface, called \ui{}, in the hope that it will facilitate research in cultural NLP. \ui{} allows users to filter papers according to our taxonomy, as well as by the cultural groups they study. For example, if a user is interested in Cultural Safety and Harm Reduction in Korean, they can easily apply filters (shown in Figures \ref{fig:ui_cc_filter} and \ref{fig:ui_culture_group_filter}) to find relevant papers (shown in Figure \ref{fig:ui_filtered_papers}) and click on a paper's title to read it. In addition to providing direct, easy access to papers, \ui{} provides dynamic bar charts at the top of the page, showing the most common Cultural Capabilities, System Methods, and Dataset Creation Methods used in the set of papers that satisfy the active set of filters. It also includes similar charts for the cultural proxies used, languages studied, and regions or countries studied. To keep up with the fast-pace of research in cultural NLP, \ui{} includes a link to a Google Form that allows anyone to submit an annotation for a paper that is not currently covered. Submitted annotations will be reviewed and/or modified manually before being added to ensure that all annotations displayed are of high-quality.


\begin{figure}[t] 
    \centering
    \includegraphics[width=\linewidth]{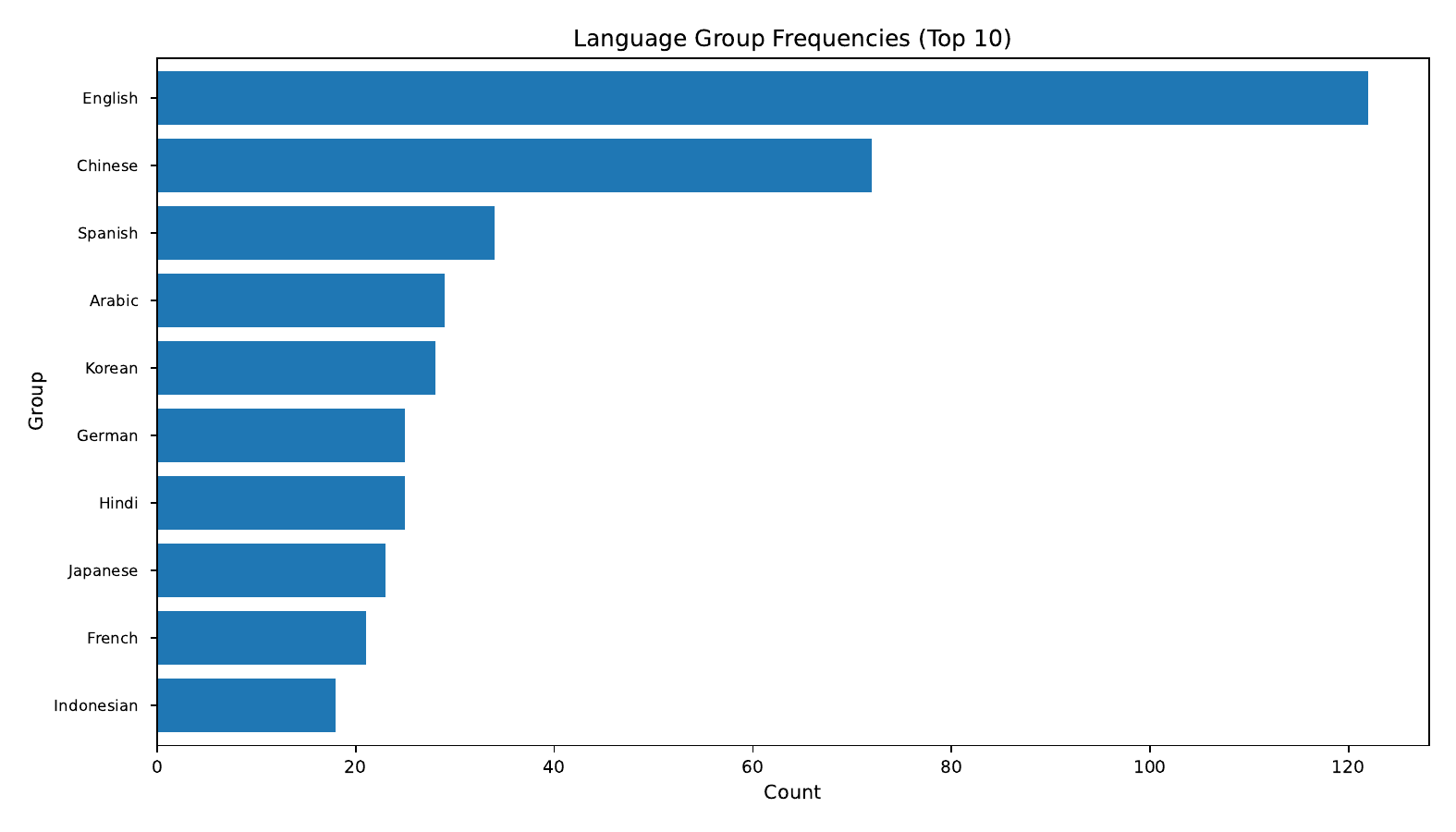}
    \caption{Bar Chart of the Frequencies of the 10 Most Commonly Studied Languages}
    \label{fig:language_counts}

    \bigskip 

    \includegraphics[width=\linewidth]{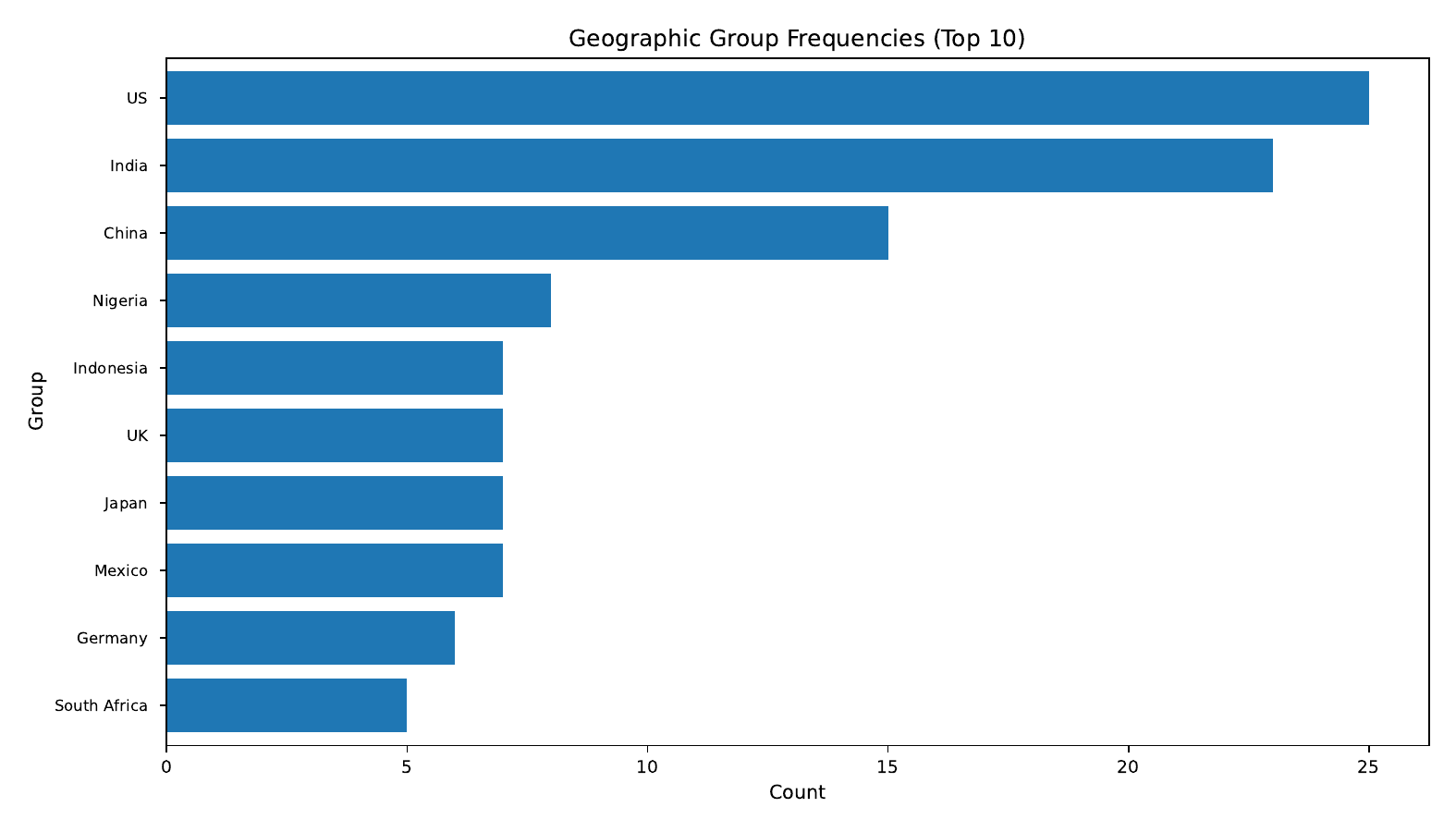}
    \caption{Bar Chart of the Frequencies of the 10 Most Commonly Studied Geographic}
    \label{fig:geographic_counts}
    
\end{figure}

\section{Cultural Proxies and Groups}

We perform a brief analysis of the cultures studied by the papers included in our survey. Figure \ref{fig:proxy_pie} breaks down the types of cultural proxies used. 72.8\% use language as the main proxy for culture, 22.2\% use a geographic proxy (Country or Geographic Region), and only 5\% use some other proxy. As discussed in \cite{zhou2025culture}, these commonly used proxies for culture can be problematic as the groups within them are often composed of multiple cultures. We further analyze the specific cultural groups included in papers that consider at most 10 different groups (235 using language and 50 using geography). Of the papers that use language, there is a very long tail distribution; over half include English and around a third include Chinese, with all but 9 languages—out of 113—being covered less than 20 times (Figure \ref{fig:language_counts}). We see a similar trend across 79 total geographic proxies, with around half including US and India, around a third including China, and all but 9 groups covered 5 or fewer times (Figure \ref{fig:geographic_counts}).

\section{List of Surveyed Papers} \label{sec:allpapers}

For compact presentation, we abbreviate cultural capability categories using acronyms in the following table of surveyed papers.
Specifically, CT $=$ Cultural Translation; 
SCA $=$ Survey-based Cultural Alignment; 
VCA $=$ Value-driven Cultural Alignment; 
CK $=$ Cultural Knowledge; 
MCK $=$ Multimodal Cultural Knowledge; 
SC $=$ Sociolinguistic Competence; 
CSHR $=$ Cultural Safety and Harm Reduction; 
CE $=$ Cultural Education; 
and CCR $=$ Cultural Computational Representation. 
These abbreviations are used only for table presentation and correspond directly to the capability categories described in the main text.

\begin{table*}[t]
\centering
\small
\setlength{\tabcolsep}{4pt}
\renewcommand{\arraystretch}{1.12}

\begin{minipage}[t]{0.485\textwidth}
\vspace{0pt}
\centering
\begin{tabular}{p{\dimexpr\linewidth-0.4\linewidth-2\tabcolsep\relax} p{0.4\linewidth}}
\hline
\textbf{Paper} & \textbf{Culture capability} \\
\hline
\citet{wang-etal-2024-seaeval} & CK, CT, VCA \\
\hline
\citet{singh-etal-2025-global} & CK, VCA \\
\hline
\citet{khanuja-etal-2020-gluecos} & SC \\
\hline
\citet{joshi-etal-2025-cultureguard} & CSHR \\
\hline
\citet{10.1145/3627673.3679768} & CCR, SC \\
\hline
\citet{tao_cultural_2024} & SCA \\
\hline
\citet{nayak-etal-2024-benchmarking} & MCK \\
\hline
\citet{xu-etal-2024-exploring-multilingual} & VCA, CCR \\
\hline
\citet{arora-etal-2023-probing} & VCA, CCR \\
\hline
\citet{cao-etal-2023-assessing} & SCA \\
\hline
\citet{conia-etal-2024-towards} & CT \\
\hline
\citet{sun-etal-2021-cross} & SC \\
\hline
\citet{alkhamissi-etal-2024-investigating} & SCA \\
\hline
\citet{masoud-etal-2025-cultural} & SCA \\
\hline
\citet{jinnai-2024-cross} & SC \\
\hline
\citet{xu-etal-2025-self} & SCA \\
\hline
\citet{ki-etal-2025-multiple} & VCA \\
\hline
\citet{sun-etal-2024-toward} & CK \\
\hline
\citet{havaldar-etal-2023-multilingual} & VCA \\
\hline
\citet{ahmad-etal-2024-generative} & VCA \\
\hline
\citet{bui-etal-2025-multi3hate} & CSHR \\
\hline
\citet{bhatia-etal-2024-local} & MCK \\
\hline
\citet{hale-etal-2025-kodis} & VCA, SC \\
\hline
\citet{liu-etal-2024-multilingual} & CK, SC \\
\hline
\citet{schneider-sitaram-2024-m5} & MCK \\
\hline
\citet{cahyawijaya-etal-2025-crowdsource} & MCK \\
\hline
\citet{tay-etal-2020-rather} & SCA \\
\hline
\citet{zheng-etal-2022-stanceosaurus} & VCA \\
\hline
\citet{mohamed-etal-2022-artelingo} & MCK \\
\hline
\citet{anegundi-etal-2022-modelling} & CSHR, CCR \\
\hline
\citet{jha-etal-2023-seegull} & CSHR, CCR \\
\hline
\citet{akinade-etal-2023-varepsilon} & CT \\
\hline
\citet{das-etal-2023-toward} & CSHR \\
\hline
\citet{bauer-etal-2023-social} & VCA \\
\hline
\citet{mukherjee-etal-2023-global} & CSHR \\
\hline
\citet{shaikh-etal-2023-modeling} & SC \\
\hline
\citet{palta-rudinger-2023-fork} & CK \\
\hline
\citet{hu-etal-2023-multi-3} & VCA \\
\hline
\citet{choenni-etal-2024-echoes} & CCR \\
\hline
\citet{naous-etal-2024-beer} & VCA \\
\hline
\citet{urailertprasert-etal-2024-sea} & MCK \\
\hline
\citet{prieto-etal-2024-translation} & CT \\
\hline
\citet{wang-etal-2024-cdeval} & VCA \\
\hline
\citet{zhou-etal-2024-large} & VCA \\
\hline
\citet{hu-etal-2024-bridging} & CT \\
\hline
\citet{khanuja-etal-2024-image} & MCK \\
\hline
\citet{wuraola-etal-2024-understanding} & SC \\
\hline
\citet{li-etal-2024-foodieqa} & MCK \\
\hline
\citet{putri-etal-2024-llm} & CK \\
\hline
\citet{mohamed-etal-2024-culture} & MCK \\
\hline
\citet{giuliani-etal-2024-cava} & CK \\
\hline
\citet{li-etal-2024-schema} & VCA \\
\hline
\citet{cao-etal-2024-bridging} & VCA \\
\hline
\citet{zhan-etal-2024-renovi} & VCA, SC \\
\hline
\citet{hsieh-etal-2024-twbias} & CSHR \\
\hline
\citet{white-etal-2024-communicate} & SC \\
\hline
\citet{yao-etal-2024-benchmarking} & CT \\
\hline
\citet{bhatt-diaz-2024-extrinsic} & VCA \\
\hline
\hline
\end{tabular}

\hfill\emph{Table continues}
\end{minipage}
\hfill
\begin{minipage}[t]{0.485\textwidth}
\vspace{0pt}
\centering
\begin{tabular}{p{\dimexpr\linewidth-0.4\linewidth-2\tabcolsep\relax} p{0.4\linewidth}}
\hline
\textbf{Paper} & \textbf{Culture capability} \\
\hline
\citet{cadotte-etal-2024-machine} & CT \\
\hline
\citet{kim-etal-2024-click} & CK \\
\hline
\citet{eo-etal-2024-detecting} & CSHR, VCA \\
\hline
\citet{zhao-etal-2024-worldvaluesbench} & SCA, CK \\
\hline
\citet{fort-etal-2024-stereotypical} & CSHR \\
\hline
\citet{havaldar-etal-2024-building} & VCA \\
\hline
\citet{lee-etal-2024-exploring-cross} & CSHR \\
\hline
\citet{abdelkadir-etal-2024-diverse} & SC \\
\hline
\citet{wang-etal-2024-kulture} & CK \\
\hline
\citet{haberland-etal-2024-italian} & CT \\
\hline
\citet{tonneau-etal-2024-languages} & CSHR \\
\hline
\citet{yakhni-chehab-2025-llms} & CT \\
\hline
\citet{arora-etal-2025-calmqa} & CK \\
\hline
\citet{belay-etal-2025-culemo} & CK \\
\hline
\citet{park-etal-2025-evaluating} & MCK \\
\hline
\citet{chiu-etal-2025-culturalbench} & CK \\
\hline
\citet{yang-etal-2025-cultural} & SC \\
\hline
\citet{havaldar-etal-2025-towards} & CT \\
\hline
\citet{wu-etal-2025-socialcc} & CK \\
\hline
\citet{zhang-etal-2025-cross} & CK \\
\hline
\citet{kim-kim-2025-dual} & CSHR, CK \\
\hline
\citet{ignat-etal-2025-inspaired} & SC \\
\hline
\citet{park-etal-2025-llm} & CSHR \\
\hline
\citet{rai-etal-2025-cross} & SC \\
\hline
\citet{kim-etal-2025-tom} & CSHR \\
\hline
\citet{kim-etal-2025-representing} & VCA \\
\hline
\citet{mousi-etal-2025-aradice} & CK, SC \\
\hline
\citet{pandey-etal-2025-culturally} & CSHR \\
\hline
\citet{cheng-etal-2025-hkcanto} & CK \\
\hline
\citet{umbet-etal-2025-kazbench} & CK \\
\hline
\citet{qiu-etal-2025-evaluating} & VCA \\
\hline
\citet{bhatia-etal-2025-swan} & CCR \\
\hline
\citet{yadav-etal-2025-beyond} & SCA \\
\hline
\citet{vasilev-etal-2025-ruscode} & MCK \\
\hline
\citet{li-etal-2025-multilingual} & VCA, CK \\
\hline
\citet{maji-etal-2025-sanskriti} & CK \\
\hline
\citet{singh-etal-2025-graph} & CT \\
\hline
\citet{schneider-etal-2025-gimmick} & CK, MCK \\
\hline
\citet{hasan-etal-2025-nativqa} & CK \\
\hline
\citet{kim-lee-2025-nunchi} & CK \\
\hline
\citet{zhang-etal-2025-creating} & MCK \\
\hline
\citet{ghaboura-etal-2025-time} & MCK \\
\hline
\citet{reshetnikov-marinescu-2025-caption} & MCK \\
\hline
\citet{anik-etal-2025-preserving} & CT \\
\hline
\citet{onohara-etal-2025-jmmmu} & CK, MCK \\
\hline
\citet{seveso-etal-2025-italic} & CK \\
\hline
\citet{bai-etal-2025-power} & CK \\
\hline
\citet{winata-etal-2025-worldcuisines} & MCK \\
\hline
\citet{naous-xu-2025-origin} & CK \\
\hline
\citet{saha-etal-2025-reading} & SC \\
\hline
\citet{berger-ponti-2025-cross} & MCK \\
\hline
\citet{jeong-etal-2025-culture} & MCK \\
\hline
\citet{thorunn-arnardottir-etal-2025-wikiqa} & CK \\
\hline
\citet{ventura-etal-2025-navigating} & CK, MCK \\
\hline
\citet{paniv-etal-2025-benchmarking} & MCK \\
\hline
\citet{ringel-etal-2019-cross} & SC \\
\hline
\citet{lin-etal-2018-mining} & SC \\
\hline
\citet{gutierrez-etal-2016-detecting} & VCA, SC \\
\hline
\hline
\end{tabular}

\hfill\emph{Table continues}
\end{minipage}

\end{table*}

\begin{table*}[t]
\centering
\small
\setlength{\tabcolsep}{4pt}
\renewcommand{\arraystretch}{1.12}

\begin{minipage}[t]{0.485\textwidth}
\vspace{0pt}
\centering
\begin{tabular}{p{\dimexpr\linewidth-0.4\linewidth-2\tabcolsep\relax} p{0.4\linewidth}}
\hline
\textbf{Paper} & \textbf{Culture capability} \\
\hline
\citet{sheng-etal-2016-dataset} & MCK \\
\hline
\citet{wilson-etal-2016-disentangling} & SC, VCA \\
\hline
\citet{friedman-etal-2019-relating} & CSHR, CCR \\
\hline
\citet{maji-etal-2025-drishtikon} & MCK \\
\hline
\citet{cuevas-etal-2025-anecdoctoring} & CSHR \\
\hline
\citet{sahoo-etal-2025-diwali} & CK \\
\hline
\citet{satar-etal-2025-seeing} & MCK \\
\hline
\citet{zhang_culturescope_2025} & CK, VCA \\
\hline
\citet{limkonchotiwat-etal-2025-wangchanthaiinstruct} & CK \\
\hline
\citet{cao-etal-2025-specializing} & SCA \\
\hline
\citet{jiang-etal-2025-language} & SCA \\
\hline
\citet{ramezani-xu-2023-knowledge} & VCA \\
\hline
\citet{cahyawijaya-etal-2025-high} & CCR \\
\hline
\citet{yin_safeworld_2024} & CSHR \\
\hline
\citet{li_culturepark_2024} & CE, CSHR, VCA \\
\hline
\citet{ziems-etal-2025-culture} & VCA, CK \\
\hline
\citet{epure-etal-2020-modeling} & CCR \\
\hline
\citet{hahm-etal-2020-crowdsourcing} & SC \\
\hline
\citet{yin-etal-2021-broaden} & MCK \\
\hline
\citet{milbauer-etal-2021-aligning} & CCR \\
\hline
\citet{liu-etal-2021-visually} & MCK \\
\hline
\citet{ghosh-etal-2021-detecting} & CSHR \\
\hline
\citet{kiesel-etal-2022-identifying} & SCA, VCA \\
\hline
\citet{kim-etal-2022-kochet} & CK \\
\hline
\citet{yin-etal-2022-geomlama} & CK \\
\hline
\citet{thai-etal-2022-exploring} & CT \\
\hline
\citet{jeong-etal-2022-kold} & CSHR \\
\hline
\citet{maronikolakis-etal-2022-listening} & CSHR \\
\hline
\citet{nadejde-etal-2022-cocoa} & CT, SC \\
\hline
\citet{sharma-etal-2022-disarm} & CSHR \\
\hline
\citet{srinivasan-choi-2022-tydip} & CSHR \\
\hline
\citet{herrera-etal-2022-tweettaglish} & SC \\
\hline
\citet{chandran-nair-etal-2022-indoukc} & CCR \\
\hline
\citet{mortensen-etal-2022-hmong} & CCR \\
\hline
\citet{deshpande-etal-2022-stereokg} & CSHR \\
\hline
\citet{ziems-etal-2023-normbank} & CCR \\
\hline
\citet{lee-etal-2023-kosbi} & CSHR \\
\hline
\citet{alshahrani-etal-2023-performance} & CCR \\
\hline
\citet{li-zhang-2023-cultural} & MCK \\
\hline
\citet{ch-wang-etal-2023-sociocultural} & CCR \\
\hline
\citet{havaldar-etal-2023-comparing} & CCR \\
\hline
\citet{lahoti-etal-2023-improving} & CK, VCA \\
\hline
\citet{koto-etal-2023-large} & CK \\
\hline
\citet{fung-etal-2023-normsage} & CCR \\
\hline
\citet{keleg-magdy-2023-dlama} & CCR \\
\hline
\citet{kabra-etal-2023-multi} & CK \\
\hline
\citet{huang-yang-2023-culturally} & CCR \\
\hline
\citet{zhou-etal-2023-cultural} & CSHR \\
\hline
\citet{bang-etal-2023-enabling} & VCA, CSHR \\
\hline
\citet{wang-etal-2024-countries} & SCA \\
\hline
\citet{zhang-etal-2024-mc2} & CCR \\
\hline
\citet{alwajih-etal-2024-peacock} & MCK \\
\hline
\citet{chen-etal-2024-logogramnlp} & CT \\
\hline
\citet{casola-etal-2024-multipico} & CSHR, SC \\
\hline
\citet{nguyen-etal-2024-seallms} & CT, CSHR \\
\hline
\citet{falk-etal-2024-overview} & CCR \\
\hline
\citet{yuan-etal-2024-translation} & CT \\
\hline
\citet{pham-etal-2024-multi} & CCR \\
\hline
\hline
\end{tabular}

\hfill\emph{Table continues}
\end{minipage}
\hfill
\begin{minipage}[t]{0.485\textwidth}
\vspace{0pt}
\centering
\begin{tabular}{p{\dimexpr\linewidth-0.4\linewidth-2\tabcolsep\relax} p{0.4\linewidth}}
\hline
\textbf{Paper} & \textbf{Culture capability} \\
\hline
\citet{feng-etal-2024-teaching} & CK \\
\hline
\citet{feng-etal-2024-modular} & SCA, VCA, CCR \\
\hline
\citet{lovenia-etal-2024-seacrowd} & CT, CK, MCK, SC, CSHR, VCA \\
\hline
\citet{watts-etal-2024-pariksha} & CK \\
\hline
\citet{acquaye-etal-2024-susu} & CK \\
\hline
\citet{aakanksha-etal-2024-multilingual} & CSHR \\
\hline
\citet{hobson-etal-2024-story} & VCA \\
\hline
\citet{mostafazadeh-davani-etal-2024-d3code} & CSHR \\
\hline
\citet{dammu_they_2024} & CSHR, SC \\
\hline
\citet{deas-etal-2024-masive} & CCR \\
\hline
\citet{wu-etal-2024-transagents} & CT \\
\hline
\citet{yuan-etal-2024-measuring} & CK \\
\hline
\citet{javed-etal-2024-indicvoices} & CCR \\
\hline
\citet{lee-etal-2024-kornat} & CK, SCA \\
\hline
\citet{yu-etal-2024-cmoraleval} & CK, VCA \\
\hline
\citet{li-etal-2024-cif} & CK \\
\hline
\citet{alyafeai-etal-2024-cidar} & CK \\
\hline
\citet{kim-etal-2024-moral} & VCA \\
\hline
\citet{wei-etal-2024-ac} & CK \\
\hline
\citet{plaza-del-arco-etal-2024-divine} & CCR \\
\hline
\citet{liu-etal-2024-evaluating-moral} & VCA \\
\hline
\citet{shi-etal-2024-culturebank} & CK, CCR \\
\hline
\citet{yuksel-etal-2024-turkishmmlu} & CK \\
\hline
\citet{masala_vorbesti_2024} & CK \\
\hline
\citet{huang-xiong-2024-cbbq} & CSHR \\
\hline
\citet{ullah-etal-2024-detecting} & CSHR \\
\hline
\citet{seth-etal-2024-dosa} & CK \\
\hline
\citet{agarwal-etal-2024-ethical} & VCA \\
\hline
\citet{tonja-etal-2024-ethiollm} & CSHR, CT \\
\hline
\citet{yarlott-etal-2024-golem} & CK \\
\hline
\citet{son-etal-2024-hae} & CK \\
\hline
\citet{benkler-etal-2024-recognizing} & VCA \\
\hline
\citet{grigoreva-etal-2024-rubia} & CSHR \\
\hline
\citet{maronikolakis-etal-2024-sociocultural} & CSHR \\
\hline
\citet{lou-etal-2024-curated} & CT \\
\hline
\citet{prabhakaran-etal-2024-grasp} & CSHR \\
\hline
\citet{espana-bonet-barron-cedeno-2024-elote} & CCR \\
\hline
\citet{shen-etal-2024-understanding} & CK \\
\hline
\citet{wang-etal-2024-cmb} & CK \\
\hline
\citet{mukherjee-etal-2024-global} & CK \\
\hline
\citet{huang-etal-2024-acegpt} & CK, VCA \\
\hline
\citet{yao-etal-2024-value} & VCA \\
\hline
\citet{acharya-etal-2024-discovering} & CK \\
\hline
\citet{liu-etal-2024-optimizing} & SC, CE \\
\hline
\citet{piccirilli_volimet_2024} & SC \\
\hline
\citet{koto-etal-2024-indoculture} & CK \\
\hline
\citet{romanyshyn-etal-2024-unlp} & CK \\
\hline
\citet{kiulian-etal-2024-bytes} & CK \\
\hline
\citet{espana-bonet-etal-2024-elote} & CCR \\
\hline
\citet{li-etal-2024-uncovering} & CCR, SC \\
\hline
\citet{zhang-etal-2024-cultural} & CT \\
\hline
\citet{anastasi-etal-2024-vida} & CSHR \\
\hline
\citet{ng-etal-2024-sghatecheck} & CSHR \\
\hline
\citet{liu-etal-2025-cultural} & VCA \\
\hline
\citet{kumar-jurgens-2025-rules} & VCA \\
\hline
\citet{sadallah-etal-2025-commonsense} & CK \\
\hline
\hline
\end{tabular}

\hfill\emph{Table continues}
\end{minipage}

\end{table*}

\begin{table*}[t]
\centering
\small
\setlength{\tabcolsep}{4pt}
\renewcommand{\arraystretch}{1.12}

\begin{minipage}[t]{0.485\textwidth}
\vspace{0pt}
\centering
\begin{tabular}{p{\dimexpr\linewidth-0.4\linewidth-2\tabcolsep\relax} p{0.4\linewidth}}
\hline
\textbf{Paper} & \textbf{Culture capability} \\
\hline
\citet{fang-etal-2025-cknowedit} & CK \\
\hline
\citet{yu-etal-2025-injongo} & SC \\
\hline
\citet{liu-etal-2025-mcs} & MCK, CK \\
\hline
\citet{zhang-etal-2025-mllms} & MCK \\
\hline
\citet{togmanov-etal-2025-kazmmlu} & CK \\
\hline
\citet{yari-koto-2025-unveiling} & CK \\
\hline
\citet{bui-etal-2025-generalization} & CK \\
\hline
\citet{menis_mastromichalakis_dont_2025} & CSHR \\
\hline
\citet{ying-etal-2025-disentangling} & CK, SC \\
\hline
\citet{feng-etal-2025-culfit} & CK \\
\hline
\citet{isbarov-etal-2025-tumlu} & VCA, CK \\
\hline
\citet{shetty-etal-2025-vital} & VCA \\
\hline
\citet{zeng-etal-2025-marco} & VCA, CK \\
\hline
\citet{yerukola-etal-2025-mind} & CSHR \\
\hline
\citet{karamolegkou-etal-2025-evaluating} & MCK \\
\hline
\citet{bhatt-etal-2025-research} & SC \\
\hline
\citet{farhansyah-etal-2025-language} & SC, CT \\
\hline
\citet{liang-etal-2025-colloquial} & CT \\
\hline
\citet{chen-etal-2025-videovista} & MCK \\
\hline
\citet{nawale-etal-2025-fairi} & CSHR \\
\hline
\citet{bayramli-etal-2025-diffusion} & MCK \\
\hline
\citet{montalan-etal-2025-batayan} & CK, CSHR \\
\hline
\citet{grandury-etal-2025-la} & VCA, CK \\
\hline
\citet{alwajih-etal-2025-palm} & CK, SC \\
\hline
\citet{olaleye-etal-2025-afrocs} & SC \\
\hline
\citet{ishita-mamidi-2025-evolution} & CT \\
\hline
\citet{alhassoun-etal-2025-saudi} & CK, VCA \\
\hline
\citet{bouamor-etal-2025-capturing} & SC \\
\hline
\citet{alwajih-etal-2025-palmx} & CK \\
\hline
\citet{krsteski-etal-2025-towards} & VCA \\
\hline
\citet{kim-johnson-2025-korean} & CSHR \\
\hline
\citet{pokharel-agrawal-2025-nediom} & SC \\
\hline
\citet{altammami-2025-leveraging} & CT \\
\hline
\citet{yang_nushurescue_2025} & CT \\
\hline
\citet{shiono-etal-2025-evaluating} & MCK \\
\hline
\citet{kabir-etal-2025-break} & SCA \\
\hline
\citet{rooein_biased_2025} & CSHR, CCR \\
\hline
\citet{maji_drishtikon_2025} & MCK \\
\hline
\citet{xuan-etal-2025-mmlu} & SC \\
\hline
\citet{sadr-etal-2025-politely} & VCA \\
\hline
\citet{fu-etal-2025-chengyu} & CK \\
\hline
\citet{shen-etal-2025-mind} & VCA \\
\hline
\citet{wang-etal-2025-multilingual} & CK, VCA \\
\hline
\citet{azmi-etal-2025-indosafety} & CSHR \\
\hline
\citet{el-mekki-etal-2025-nilechat} & CT, VCA, CK \\
\hline
\citet{hu_toxicity_2025} & CSHR \\
\hline
\citet{ramezani-xu-2025-discordance} & CK \\
\hline
\citet{tanwar_you_2025} & CK \\
\hline
\citet{singh_lets_2025} & CK, MCK \\
\hline
\citet{hashmat-etal-2025-pakbbq} & CSHR \\
\hline
\citet{liu-etal-2025-synthetic} & SC \\
\hline
\citet{xie-etal-2025-large} & CK \\
\hline
\citet{yamamoto-etal-2025-bias} & CSHR \\
\hline
\citet{aji-cohn-2025-loraxbench} & CK, CT \\
\hline
\citet{mia-etal-2025-banmime} & MCK, CSHR \\
\hline
\citet{chen-etal-2025-steer} & CK \\
\hline
\citet{liu-etal-2025-beyond-demographics} & SCA \\
\hline
\citet{xu_punmemecn_2025} & MCK \\
\hline
\hline
\end{tabular}

\hfill\emph{Table continues}
\end{minipage}
\hfill
\begin{minipage}[t]{0.485\textwidth}
\vspace{0pt}
\centering
\begin{tabular}{p{\dimexpr\linewidth-0.4\linewidth-2\tabcolsep\relax} p{0.4\linewidth}}
\hline
\textbf{Paper} & \textbf{Culture capability} \\
\hline
\citet{mukherjee-etal-2025-women} & VCA \\
\hline
\citet{zhang-etal-2025-llm-driven} & MCK \\
\hline
\citet{shafique-etal-2025-culturally} & MCK \\
\hline
\citet{al-ghallabi-etal-2025-fann} & CK \\
\hline
\citet{parappan-henao-2025-learning} & CSHR \\
\hline
\citet{lavrouk-etal-2025-foundation} & MCK, CK \\
\hline
\citet{calvo-bartolome-etal-2025-discrepancy} & CK \\
\hline
\citet{chen-etal-2025-large-language-models} & CT \\
\hline
\citet{seweryn-etal-2025-pllum} & CK, CSHR, VCA \\
\hline
\citet{ma-etal-2025-scalable} & CSHR \\
\hline
\citet{nyandwi-etal-2025-grounding} & MCK \\
\hline
\citet{dai-etal-2025-word} & SCA \\
\hline
\citet{zhou-etal-2025-hanfu} & MCK \\
\hline
\citet{yang-etal-2025-mrguard} & CSHR \\
\hline
\citet{mitran-etal-2025-probing} & SCA \\
\hline
\citet{roh-etal-2025-xlqa} & CK \\
\hline
\citet{zhang-etal-2025-litransproqa} & CT \\
\hline
\citet{sitaram-etal-2025-multilingual} & CSHR \\
\hline
\citet{zhong-etal-2025-pluralistic} & VCA \\
\hline
\citet{nimo-etal-2025-africa} & CK, CSHR \\
\hline
\citet{guo-etal-2025-care} & CK \\
\hline
\citet{vasselli-etal-2025-multilingual} & SC, CT \\
\hline
\citet{pramodya-etal-2025-sinhalammlu} & CK \\
\hline
\citet{chen-etal-2025-benchmarking-llms} & CT \\
\hline
\citet{hong-etal-2025-normgenesis} & SC, VCA \\
\hline
\citet{hsieh-etal-2025-taiwanvqa} & MCK \\
\hline
\citet{r-etal-2025-field} & SC \\
\hline
\citet{hosseinbeigi-etal-2025-advancing} & CK \\
\hline
\citet{korre-etal-2025-untangling} & CSHR \\
\hline
\citet{saffari-etal-2025-introduce} & VCA, CSHR \\
\hline
\citet{tapo-etal-2025-gaife} & CE \\
\hline
\citet{wang-etal-2025-proverbs} & CT \\
\hline
\citet{lan-etal-2025-mcbe} & CSHR \\
\hline
\citet{dwivedi-etal-2025-eticor} & VCA \\
\hline
\citet{sasu-etal-2025-akan} & SC \\
\hline
\citet{susanto-etal-2025-sea} & CK, VCA, SC \\
\hline
\citet{tsutsumi-jinnai-2025-large} & CK \\
\hline
\citet{hosseinbeigi-etal-2025-matina-culturally} & VCA, CK, CT \\
\hline
\citet{he-etal-2025-chumor} & CK \\
\hline
\citet{das-etal-2025-yinyang} & MCK, CSHR \\
\hline
\citet{nahin-etal-2025-titullms} & CK \\
\hline
\citet{bi-etal-2025-dongbamie} & MCK \\
\hline
\citet{zou-etal-2025-llms} & CT \\
\hline
\citet{wang-etal-2025-dlir} & CCR, CK \\
\hline
\citet{tan-etal-2025-benchmark} & CT \\
\hline
\citet{zhang-etal-2025-interesting} & SC \\
\hline
\citet{mor-lan-etal-2025-hebid} & CCR \\
\hline
\citet{mubarak-etal-2025-arasafe} & CSHR \\
\hline
\citet{zhang-etal-2025-culturesynth} & CK \\
\hline
\citet{chae-etal-2025-assessing} & CK, VCA, CSHR \\
\hline
\citet{ma-etal-2025-enhancing} & SC \\
\hline
\citet{trager-etal-2025-mftcxplain} & CSHR \\
\hline
\citet{gupta-etal-2025-endive} & SC \\
\hline
\citet{dey-etal-2025-llms} & SCA \\
\hline
\citet{nayak-etal-2025-culturalframes} & MCK \\
\hline
\citet{villa-cueva-etal-2025-cammt} & CT \\
\hline
\citet{wibowo-etal-2024-copal} & CK \\
\hline
\citet{alwajih-etal-2025-pearl} & MCK \\
\hline
\citet{zhao-etal-2025-makieval} & CK \\
\hline
\hline
\end{tabular}

\hfill\emph{Table continues}
\end{minipage}

\end{table*}

\begin{table*}[t]
\small
\setlength{\tabcolsep}{4pt}
\renewcommand{\arraystretch}{1.12}

\begin{minipage}[t]{0.485\textwidth}
\vspace{0pt}
\centering
\begin{tabular}{p{\dimexpr\linewidth-0.4\linewidth-2\tabcolsep\relax} p{0.4\linewidth}}
\hline
\textbf{Paper} & \textbf{Culture capability} \\
\hline
\citet{malik_are_2025} & SC, VCA \\
\hline
\citet{caplan_splits_2025} & SC \\
\hline
\citet{mohammadi-etal-2025-large} & SCA \\
\hline
\citet{bennie-etal-2025-panda} & CSHR \\
\hline
\citet{rachamalla-etal-2025-pragyaan} & CK, CSHR, SC \\
\hline
\citet{liao-etal-2025-culture} & CT \\
\hline
\citet{pujari-goldwasser-2025-llm} & VCA, CCR \\
\hline
\citet{muhammad-etal-2025-afrihate} & CSHR \\
\hline
\citet{madhusudan_fine-tuned_2025} & CCR \\
\hline
\citet{leteno-etal-2025-histoiresmorales} & VCA \\
\hline
\citet{son-etal-2025-kmmlu} & CK \\
\hline
\citet{ashraf-etal-2025-arabic} & CSHR \\
\hline
\citet{khanuja-etal-2025-towards} & MCK \\
\hline
\citet{banerjee_navigating_2025} & CSHR \\
\hline
\citet{miehling-etal-2025-evaluating} & VCA \\
\hline
\citet{nikandrou-etal-2025-crope} & MCK \\
\hline
\citet{pham-etal-2025-cultureinstruct} & VCA, CSHR \\
\hline
\citet{zhou-etal-2025-mapo} & CK \\
\hline
\citet{verma-etal-2025-milu} & CK \\
\hline
\citet{rai-etal-2025-social} & VCA \\
\hline
\citet{mitchell-etal-2025-shades} & CSHR \\
\hline
\citet{tapo-etal-2025-bayelemabaga} & CT \\
\hline
\citet{magdy-etal-2025-jawaher} & CK \\
\hline
\citet{moosavi-monazzah-etal-2025-percul} & CK \\
\hline
\citet{abuhaija-etal-2025-nakba} & CCR \\
\hline
\citet{volker-etal-2025-salt} & CT \\
\hline
\citet{conia-etal-2025-semeval} & CT \\
\hline
\citet{park-etal-2025-polite} & VCA \\
\hline
\citet{srirag-etal-2025-evaluating} & SC \\
\hline
\citet{sahoo-etal-2024-indibias} & CSHR \\
\hline
\citet{ozbal-etal-2016-prometheus} & VCA, SC \\
\hline
\hline
\end{tabular}
\end{minipage}
\end{table*}

\end{document}